\documentclass[10pt,twocolumn,letterpaper]{article}

\usepackage{iccv}
\usepackage{times}
\usepackage{epsfig}
\usepackage{graphicx}
\usepackage{amsmath}
\usepackage{amssymb}

\usepackage{bbm}
\usepackage{multirow}
\usepackage[numbers,sort]{natbib}
\usepackage{enumitem}
\usepackage{caption}
\usepackage{subcaption}
\usepackage{booktabs}
\usepackage{pifont}
\usepackage[accsupp]{axessibility}

\usepackage[table]{xcolor}
\usepackage{tikz}
\usepackage{comment}
\usepackage{makecell}
\usepackage{array, boldline}
\usepackage{tabularx}
\usepackage{booktabs}
\usepackage{hhline}
\usepackage{highlight}
\usepackage[pagebackref=true,breaklinks=true,letterpaper=true,colorlinks,bookmarks=false]{hyperref}
\usepackage{color}

\usepackage[pagebackref=true,breaklinks=true,letterpaper=true,colorlinks,bookmarks=false]{hyperref}

\iccvfinalcopy 

\def\iccvPaperID{5805} 

\ificcvfinal\fi

\begin{document}

\title{Label-Free Event-based Object Recognition via Joint Learning with\\ Image Reconstruction from Events}

\author{Hoonhee Cho*, Hyeonseong Kim*, Yujeong Chae, and Kuk-Jin Yoon \\
Korea Advanced Institute of Science and Technology \\
{\tt\small \{gnsgnsgml,brian617,yujeong,kjyoon\}@kaist.ac.kr}
}

\maketitle
\def\thefootnote{*}\footnotetext{The first two authors contributed equally. In alphabetical order. }\def\thefootnote{\arabic{footnote}}

\maketitle
\ificcvfinal\thispagestyle{empty}\fi

\begin{abstract}
Recognizing objects from sparse and noisy events becomes extremely difficult when paired images and category labels do not exist. In this paper, we study label-free event-based object recognition where category labels and paired images are not available.
To this end, we propose a joint formulation of object recognition and image reconstruction in a complementary manner.
Our method first reconstructs images from events and performs object recognition through Contrastive Language-Image Pre-training (CLIP), enabling better recognition through a rich context of images.
Since the category information is essential in reconstructing images, we propose category-guided attraction loss and category-agnostic repulsion loss to bridge the textual features of predicted categories and the visual features of reconstructed images using CLIP. 
Moreover, we introduce a reliable data sampling strategy and local-global reconstruction consistency to boost joint learning of two tasks. To enhance the accuracy of prediction and quality of reconstruction, we also propose a prototype-based approach using unpaired images. Extensive experiments demonstrate the superiority of our method and its extensibility for zero-shot object recognition. Our project code is available at \url{https://github.com/Chohoonhee/Ev-LaFOR}.
\end{abstract}

\begin{figure}[t] 
\centering 
\includegraphics[width=.99\columnwidth]{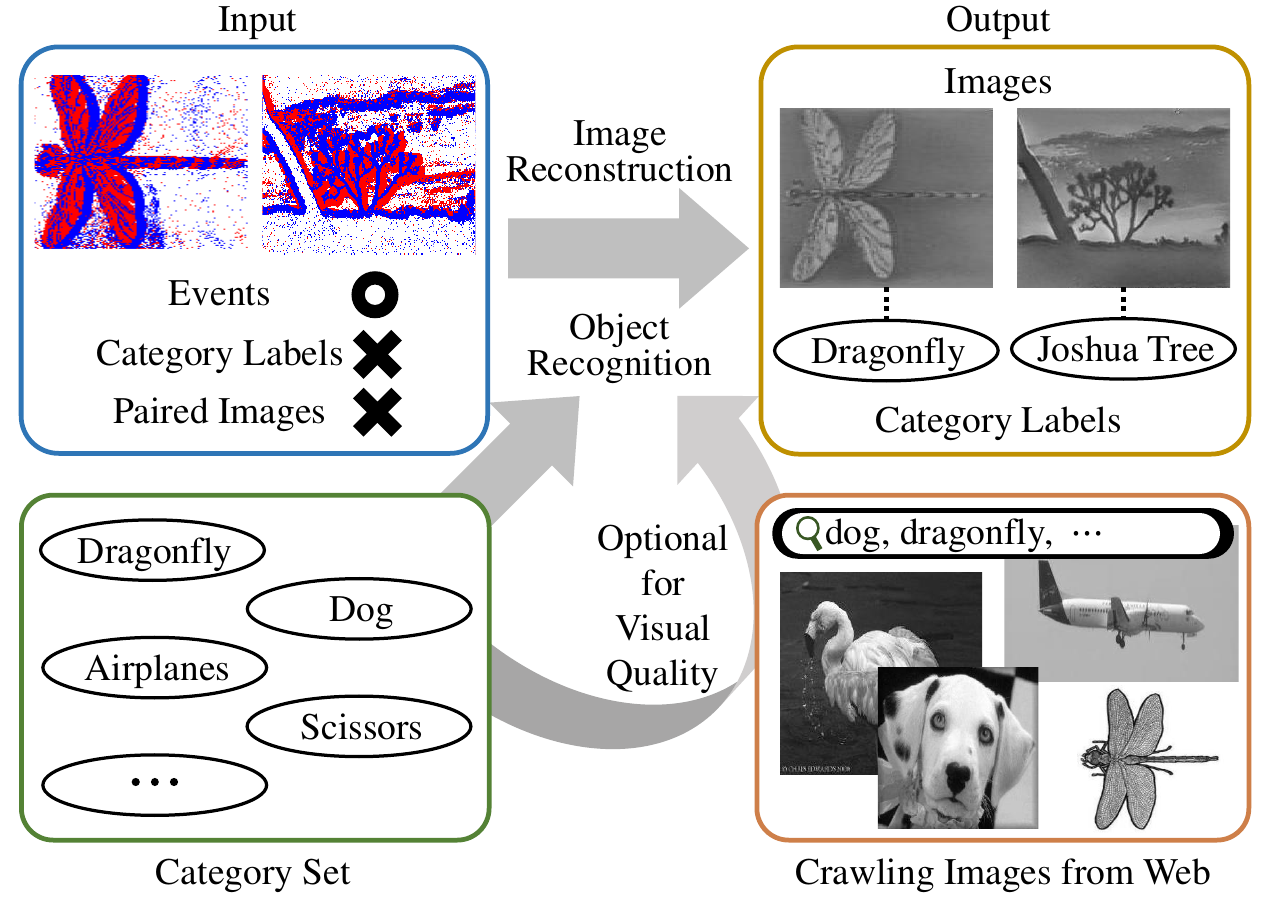} 
\caption{We tackle label-free event-based object recognition where category labels and paired images are not available. Our approach can simultaneously perform object recognition and image reconstruction through proposed joint learning framework. Optionally, our method can leverage unpaired images to enhance object recognition performance and reconstructed image quality.
}
\label{fig:teaser}
\end{figure}

\vspace{-10pt}
\section{Introduction}

Event cameras are neuromorphic vision sensors that asynchronously perceive per-pixel brightness changes with high temporal resolution. They show advantages over conventional frame-based cameras, such as high dynamic range (HDR) and low latency. However, despite the superiority of cameras, algorithms using events are still in their infancy.
Existing event-based object recognition methods~\cite{lee2016training, orchard2013spiking, orchard2015hfirst,neil2016phased,perez2013mapping,lagorce2016hots,amir2017low,sironi2018hats,messikommer2020event, cannici2020differentiable,kim2021n,klenk2022masked} have been shown that object recognition is a crucial task for event feature learning \cite{cohen2018spatial, maqueda2018event, park2016performance} and endeavors have successfully classified the object with events.
However, they rely on supervision with category labels to achieve high performance.
Furthermore, the sparsity of events poses a challenge for users to label them accurately when there are no paired images.
Several unsupervised methods \cite{lagorce2016hots, orchard2015hfirst} have been proposed for object recognition, however, their performances remain subpar.
Hence, there is a natural need for robust and effective label-free event-based object recognition methods.

In this paper, we study label-free event-based object recognition where category labels and paired images are not available as in Fig.~\ref{fig:teaser}.
To tackle this task, we focus on two following facts.
Firstly, as demonstrated in~\cite{rebecq2019high, wang2021joint}, performing object recognition on reconstructed images with spatially dense information and rich context shows better performance than solely using events.
Secondly, during reconstructing images from events, taking the categories into account can improve the quality of the reconstructed images. 
From these observations, we propose a joint learning framework of two closely-linked tasks: event-based object recognition and event-to-image reconstruction.

In our joint learning framework, we first reconstruct images from events and perform object recognition on the reconstructed images using Contrastive Language-Image Pre-training (CLIP)~\cite{radford2021learning_clip}. 
CLIP is an image-text embedding model that has been trained on a large number of image and language pairs sourced from the web.
We utilize the textual features encoded from CLIP since they are well-aligned with visual features, which is advantageous for our tasks that lack labels and images.
To this end, we propose the category-guided attraction loss that aligns the visual features from reconstructed images close to the texture features of the predicted category from CLIP. 

Although the above framework enables the joint learning of event-based object recognition and image reconstruction, using the predicted categories brings two issues. (1) Collapsing to the trivial solution, resulting in all outputs being assigned to a single category and (2) Interrupting the learning process due to unreliable~(wrong) predictions.
To alleviate these problems, we propose (1) category-agnostic repulsion loss to increase the distances between visual features for preventing collapse and (2) a reliable data sampling~(RDS) method to reduce the effects of unreliable predictions. 
For RDS, we devise two reliable indicators named posterior probability indicator~(PPI) and temporally reversed consistency indicator~(TRCI). The PPI selects reliable samples based on the probability of prediction. For additional sampling, TRCI utilizes the characteristics of events that temporally reversed ones should be categorized into the same category as the original events.
This reliable data sampling enables stable learning even with the predicted categories, leading to a substantial performance improvement in object recognition.
Furthermore, to restore the local details, we introduce the local-global reconstruction consistency.
Finally, to enhance the accuracy of recognition and visual quality of reconstructed images, we expand our method by employing the existing unpaired images. 
Extensive experiments demonstrate that our method shows outstanding results on the subset of the N-ImageNet dataset over unsupervised and image reconstruction-based approaches and even surpasses the supervised methods on the N-Caltech101 dataset without using labels and images.
In addition, we show that our framework can be extended to event-based zero-shot object recognition.

Our work presents four main contributions: (I) We propose a novel joint learning framework for event-based object recognition and image reconstruction, without requiring paired images and labels. (II) We introduce a reliable data sampling strategy and local-global reconstruction consistency that enhances label-free joint training. (III) We further introduce employing unpaired images in our framework for boosting performance. (IV) We conduct extensive experiments of our method along with zero-shot object recognition and superset categories learning, and demonstrate our method's superior performance and effectiveness.

\vspace{-2pt}
\section{Related Work}
\vspace{-2pt}
\noindent
\textbf{Event-based Object Recognition.} 
Concurrent with the success of event-based low-level tasks~\cite{cho2021eomvs, cho2022selection, cho2022event, kim2023event, sun2022event, zhu2019unsupervised, shiba2022secrets}, there has been a surge of active research focused on object recognition~\cite{kim2021n,ncaltech, serrano2015poker,li2017cifar10,bi2019graph,sironi2018hats,moeys2018pred18,hu2016dvs,cheng2019det,lungu2017live,vasudevan2020introduction,ramesh2020low}, which is considered as a fundamental task. Considering the unique features of event data, numerous works~\cite{lee2016training, orchard2013spiking, orchard2015hfirst,neil2016phased,perez2013mapping,lagorce2016hots,amir2017low,sironi2018hats,messikommer2020event, cannici2020differentiable,kim2021n,klenk2022masked} have presented models and representations that can effectively extract significant patterns.
However, these approaches require labels for each event data for supervised learning, and since event data cannot be crawled from the web like images, the labeling cost of event data is extremely high. On the other hand, our approach enables the learning of event-based object recognition and image reconstruction simultaneously, without requiring category labels and paired images.

\noindent
\textbf{Event-based Image Reconstruction} is a well-known topic in the field of event-based vision~\cite{kim2008simultaneous, kim2016real, bardow2016simultaneous, reinbacher2016real, scheerlinck2019continuous, cook2011interacting,gehrig2018asynchronous,wang2021dual}. Recent researches have been utilizing deep learning to generate realistic outcomes~\cite{rebecq2019high, e2vid, scheerlinck2020fast, weng2021event}. However, these supervised approaches require the paired images of each event stream to be accurately aligned and synchronized in terms of pixels and time.

Another line of approach, such as Wang~\etal~\cite{wang2019event} and Pini~\etal~\cite{pini2018learn}, utilized generative adversarial networks (GANs) to reconstruct intensity with unpaired images that are not related to the events of the scene. However, GANs are prone to be sensitive to various parameters, and when images that are significantly irrelevant to objects are utilized, the quality of the reconstructed images from events is notably low. An alternative solution is a self-supervised method~\cite{paredes2021back} that utilizes photometric consistency, but this approach requires a continuous sequence that spans an adequately long period. 

Differently, our framework can reconstruct the intensity of event data without images, very long event sequence and GANs. We only exploit the category prompts of a bundle of event data, not the class label of each event data.

\begin{figure*}[t] 
\centering 
\vspace{-2pt}
\includegraphics[width=.99\linewidth]{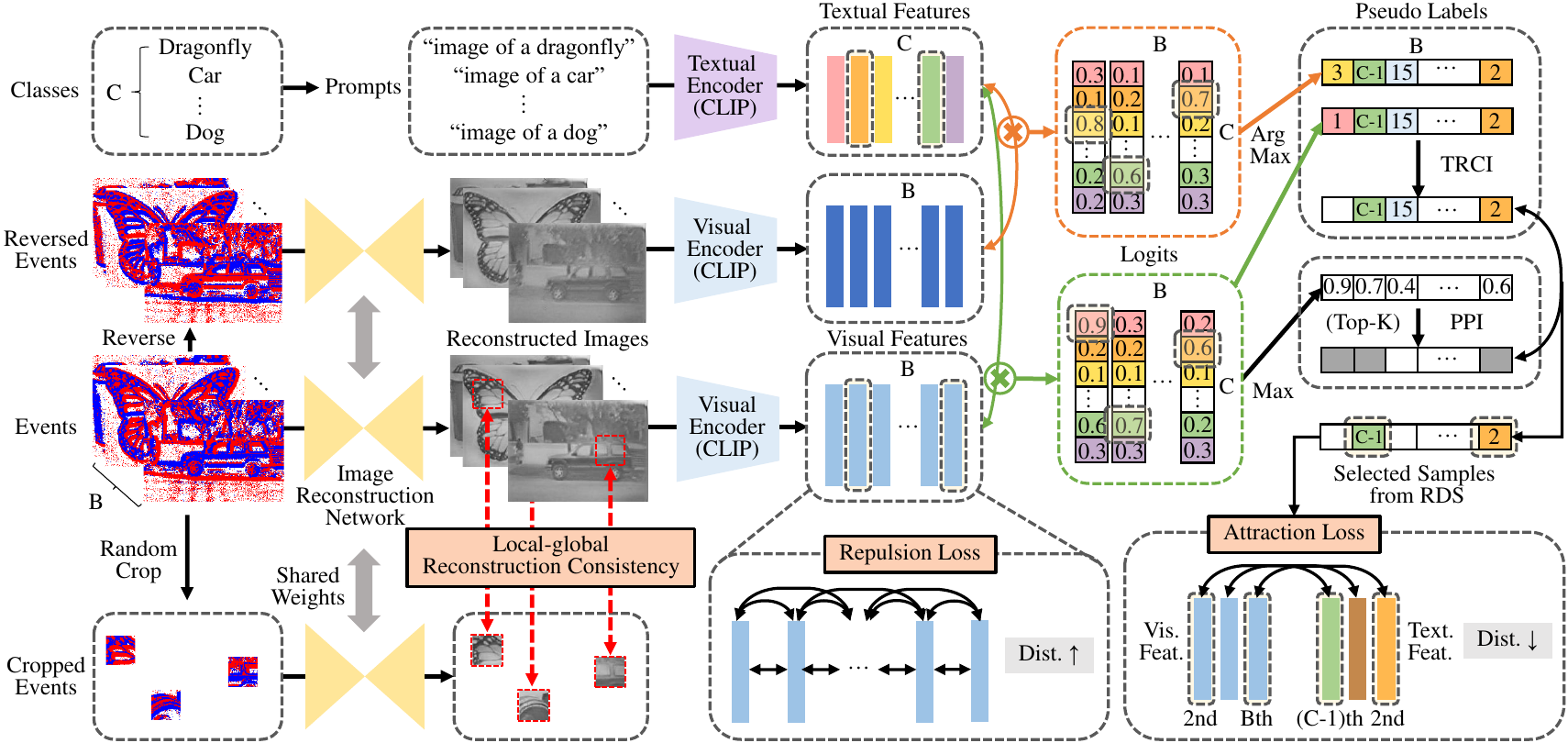} 
\vspace{-5pt}
\caption{Overall joint learning framework of event-based object recognition with event-to-image reconstruction.
Our approach involves CLIP's textual and visual encoders that derive textual and visual features, respectively, from given class sets and events. Textual features are obtained from category prompts, while visual features are obtained from reconstructed images. Then, logits and corresponding pseudo labels are generated by computing the cosine similarity between visual and textual features. Lastly, our framework employs pseudo labels to facilitate joint learning using category-agnostic repulsion loss and category-guided attraction loss. To mitigate the impact of unreliable pseudo labels and boost performance, we introduce a reliable data sampling strategy and local-global reconstruction consistency.
}
\label{fig:framework}
\vspace{-10pt}
\end{figure*}

\noindent
\textbf{Label-free Learning} aims to solve the end-task only with unlabeled data, and has been actively explored~\cite{wang2021evdistill,cho2023learning}.
HOTS~\cite{lagorce2016hots} proposed a category-free method for recognizing objects based on events and SSL-E2VID~\cite{paredes2021back} proposed a self-supervised learning method for reconstructing images from only events.
However, their performances fall short of supervised learning methods with significant performance gaps.
Inspired by the success of using the pre-trained model~\cite{tewel2022zerocap, chefer2022image, patashnik2021styleclip,chen2023clip2scene, sanghi2022clip}, we adopt CLIP~\cite{radford2021learning_clip} for label-free event-based object recognition. 
We propose a novel joint learning framework that uses visual and textual knowledge to achieve superior performance in event-to-image reconstruction and event-based object recognition without category labels and paired images. To the best of our knowledge, we are the first to incorporate textual knowledge into event vision research, particularly in the context of label-free tasks.

\vspace{-2pt}
\section{Method}
\vspace{-2pt}
In Sec.~\ref{sec:joint_form}, we first introduce our joint formulation of event-to-image reconstruction and object recognition for label-efficient learning.
In Sec.~\ref{sec:rds}, we describe our reliable data sampling~(RDS) strategy for boosting joint learning performance by filtering the unreliable data samples during training.
Sec.~\ref{sec:local-global} introduces the local-global reconstruction consistency for spatially regularizing the reconstructed images, which ensures capturing the local details.
In Sec.~\ref{sec:employ_unpair}, we propose to further employ the unpaired images for better recognition performance and reconstructed image quality.

\vspace{-2pt}
\subsection{Joint Formulation}
\vspace{-2pt}
\label{sec:joint_form}
We propose a joint formulation of event-to-image reconstruction and object recognition to perform event-based (zero-shot) object recognition without using labels and paired images of event data.
Our joint formulation is based on the fact that 1) image reconstruction can be helpful in predicting categories from events, and 2) category information is important in reconstructing images from events.
Therefore, we propose 1) the process of performing event-to-image reconstruction and CLIP-driven object recognition from the reconstructed images and 2) category-guided attraction loss and category-agnostic repulsion loss for learning this process without labels and paired images.
The overall framework is shown in Fig.~\ref{fig:framework}.
Note that our joint learning framework can reconstruct intensity images from events without using any images.

\noindent\textbf{Event-to-Image Reconstruction.}
For the joint formulation, we first reconstruct the intensity image $\mathcal{I}$ from the event stream $\mathcal{E} = (u_i, t_i, p_i)_{i=1}^{N}$, where $u_i$, $t_i$, and $p_i$ denote triggered pixel location, time, and polarity of each event, and $N$ is the temporal length of the event stream.
We convert the event stream into event spiking tensor~(EST)~\cite{gehrig2019end}, which is then fed into the image reconstruction network $\mathcal{G}$ to obtain the intensity image.
The image reconstruction process is formulated as follows:
\vspace{-3pt}
\begin{equation}
\label{eq:recon}
    \mathcal{I} = \mathcal{G}(\mathrm{EST}(\mathcal{E})).
\vspace{-3pt}
\end{equation}

\noindent\textbf{CLIP-driven Object Recognition.}
We perform object recognition on top of the reconstructed images instead of raw event data.
The advantage of using reconstructed images rather than using event data directly for object recognition is that better performance can be achieved by using a network trained on rich image data~\cite{rebecq2019high,wang2021joint}.
In particular, we use CLIP~\cite{radford2021learning_clip} for object recognition, which also enables zero-shot object recognition.
For an event dataset of $C$ classes, the category prompts are constructed by placing each category name into a pre-defined template, ``image of a [CLASS]."
Then, CLIP's textual encoder encodes category prompts to obtain the $D$-dimensional textual features $F = (f_i)_{i=1}^{C} \in \mathbb{R}^{C \times D}$.

Meantime, the reconstructed image $\mathcal{I}$ is encoded by CLIP's visual encoder to a visual feature $v \in \mathbb{R}^{1 \times D}$ and the class probability $p \in \mathbb{R}^{1 \times C}$ for the object recognition are obtained as,

\begin{equation}
\label{eq:recognition}
    p_i = \mathrm{softmax}_{i}(vF^\top),
\end{equation}
where $p_i$ and $\mathrm{softmax}_{i}(\cdot)$ denote the probability for category $i$ and the softmax function.
By taking the category of maximum probability, we can predict the category $c$ for the event data $\mathcal{E}$ as,
\begin{equation}
\label{eq:prediction}
    c = \underset{i}{\mathrm{argmax}}(p_i).
\end{equation}

\noindent\textbf{Category-guided Attraction Loss.}
The challenging part of our joint formulation is to train the reconstruction network $\mathcal{G}$ without using paired images of event data.
Once we reconstruct the intensity images from the event stream~(Eq.~\ref{eq:recon}), we can perform object recognition using CLIP~(Eq.~\ref{eq:recognition}).
Motivated by the fact that category information is important in reconstructing images from events, we propose category-guided attraction loss to train the reconstruction network.
The proposed category-guided attraction loss brings the visual feature $v$ encoded from the reconstructed image $\mathcal{I}$ closer to the textual feature of the corresponding category, guiding the model to reconstruct the image corresponding to the category.
Since category labels are not available for the event data, we use the predicted categories as pseudo labels.
For event data with a batch size of $B$, the category-guided attraction loss $\mathcal{L}_{att}$ is formulated as InfoNCE loss~\cite{infonce}:
\vspace{-5pt}
\begin{equation}
\label{eq:att_loss}
    \mathcal{L}_{att} = -\sum_{i=1}^{B}{\log{\frac{\exp(v_i f_{c_i}^\top)}{\sum_{j=1}^{B}{\exp(v_i f_{c_j}^\top)}}}}, 
\end{equation}
where $v_i$, $c_i$, and $f_j$ denote the visual feature and predicted category of the $i^{\mathrm{th}}$ event data, and the textual feature of category $j$.
Minimizing $\mathcal{L}_{att}$ enables the image reconstruction network to generate images of the corresponding category label, while object recognition benefits from the reconstructed images.

\noindent\textbf{Category-agnostic Repulsion Loss.}
Although $\mathcal{L}_{att}$ enables joint learning, the use of pseudo labels may cause collapsing, resulting in all outputs being assigned to a single category.
Therefore, we introduce a category-agnostic repulsion loss to prevent collapsing by increasing the distances between visual features. 
To be specific, the category-agnostic repulsion loss $\mathcal{L}_{rep}$ is applied to the set of visual features as follows:
\vspace{-5pt}
\begin{equation}
\label{eq:attloss}
    \mathcal{L}_{rep} = \sum_{i=1}^{B}{\log(1+{{\sum_{j=1, j \neq i}^{B}{\exp(v_i v_j^\top)}})}}.
\end{equation}
With $\mathcal{L}_{att}$ and $\mathcal{L}_{rep}$, we can successfully formulate our joint learning framework, enabling simultaneous event-to-image reconstruction and (zero-shot)~object recognition.

\vspace{-2pt}
\subsection{Reliable Data Sampling}
\vspace{-2pt}
\label{sec:rds}
As $\mathcal{L}_{att}$ relies on pseudo labels, the reliability of predictions may affect the overall training procedure and performance.
To reduce the effects of unreliable predictions, we devise two reliability indicators named posterior probability indicator~(PPI) and temporally reversed consistency indicator~(TRCI).
Based on the resultant reliability of predictions, we sample data with reliable predictions for $\mathcal{L}_{att}$.
The remaining unreliable data is excluded from calculating $\mathcal{L}_{att}$ and used only for $\mathcal{L}_{rep}$.

\noindent\textbf{Posterior Probability Indicator.}
Given the reconstructed image $\mathcal{I}$ and the predicted category $c$, the posterior probability $p_c = \mathbf{p}(c|\mathcal{I})$ indicates the model confidence for category $c$.
Thus, we use the posterior probability as an indicator of the reliability of the prediction.
In particular, we select $K$ data samples corresponding to the top-$K$ posterior probabilities among $B$ data samples in the mini-batch.
The set of selected sample indices from PPI, $S_{\mathrm{PPI}}$ is defined as,
\begin{equation}
\label{eq:ppi}
    S_{\mathrm{PPI}} = [\underset{i}{\mathrm{argsort}} (\mathbf{p}(c_i | \mathcal{I}_i))]_K,
\vspace{-5pt}
\end{equation}
where $\mathcal{I}_i$ and $[\cdot]_K$ denote the reconstructed image of the $i^\mathrm{th}$ event data in the mini-batch and the set of the first $K$ items.
The $\mathrm{argsort}(\cdot)$ function returns indices that would sort an input in descending order.

\noindent\textbf{Temporally Reversed Consistency Indicator.}
Considering the temporal aspect of event data, we propose an additional reliability indicator based on the consistency between predictions from the original events and the temporally reversed events.
From the fact that the temporally reversed events $\mathcal{E}^R = (u_i, \max_j(t_j)-t_i, p_i)_{i=N}^{1}$ should be categorized into the same category as the original events $\mathcal{E}$, we presume that the consistent model prediction would indicate the reliability of the data sample.
To be specific, the set of selected sample indices from TRCI, $S_{\mathrm{TRCI}}$ is defined as follows:
\vspace{-5pt}
\begin{equation}
\label{eq:trci}
    S_{\mathrm{TRCI}} = \{i | c_i = c^R_i,\, i=1,2,\cdots,B\},
\end{equation}
where $c^R_i$ denotes a predicted category of $i^{\mathrm{th}}$ reversed event data $\mathcal{E}^R_i$ in the mini-batch.

Our final set of selected sample indices $S_\mathrm{RDS}$ for reliable data sampling~(RDS) is the intersection of $S_\mathrm{PPI}$ and $S_\mathrm{TRCI}$:
\begin{equation}
\label{eq:rds}
    S_{\mathrm{RDS}} = S_\mathrm{PPI} \cap S_\mathrm{TRCI}.
\end{equation}
Then, the modified attraction loss with RDS is as follows:
\begin{equation}
\label{eq:mod_att_loss}
    \mathcal{L}_{att} = -\sum_{i \in S_{\mathrm{RDS}}}{\log{\frac{\exp(v_i f_{c_i}^\top)}{\sum_{j \in S_{\mathrm{RDS}}}{\exp(v_i f_{c_j}^\top)}}}}, 
\end{equation}

\vspace{-2pt}
\subsection{Local-global Reconstruction Consistency}
\vspace{-2pt}
\label{sec:local-global}
Along with $\mathcal{L}_{att}$ and $\mathcal{L}_{rep}$, we further introduce the local-global reconstruction consistency for spatially regularizing the reconstructed images in a self-supervised manner.
With random crop function $\mathcal{H}$, we first crop the EST of original events $\mathrm{EST}(\mathcal{E})$ to obtain the cropped tensor $\mathcal{H}(\mathrm{EST}(\mathcal{E}))$ for the local input.
Then, the local-global reconstruction consistency loss $L_{con}$ is defined as,
\vspace{-2pt}
\begin{equation}
\label{eq:rds}
    \mathcal{L}_{con} = \| \mathcal{G}(\mathcal{H}(\mathrm{EST}(\mathcal{E}))) - \mathcal{H}(\mathcal{G}(\mathrm{EST}(\mathcal{E}))) \|_1.
\vspace{-2pt}
\end{equation}
Compared to $\mathcal{L}_{att}$ and $\mathcal{L}_{rep}$ that work on visual features, $L_{con}$ is applied directly to the reconstructed images and considered as a spatial regularization for the reconstruction network $\mathcal{G}$.
In addition, matching the reconstructed images from local and global inputs enables the network to capture the local details, improving the reconstruction quality.

The total loss function for the joint learning framework is as follows:
\vspace{-2pt}
\begin{equation}
\label{eq:total_loss}
    \mathcal{L}_{total} = \lambda_1 \mathcal{L}_{att} + \lambda_2 \mathcal{L}_{rep} + \lambda_3 \mathcal{L}_{con},
\vspace{-2pt}
\end{equation}
where $\lambda_1$, $\lambda_2$, and $\lambda_3$ are weights for $\mathcal{L}_{att}$, $\mathcal{L}_{rep}$, and $\mathcal{L}_{con}$.

\vspace{-2pt}
\subsection{Employing Unpaired Images}
\begin{figure}[t]
\centering 
\includegraphics[width=.99\columnwidth]{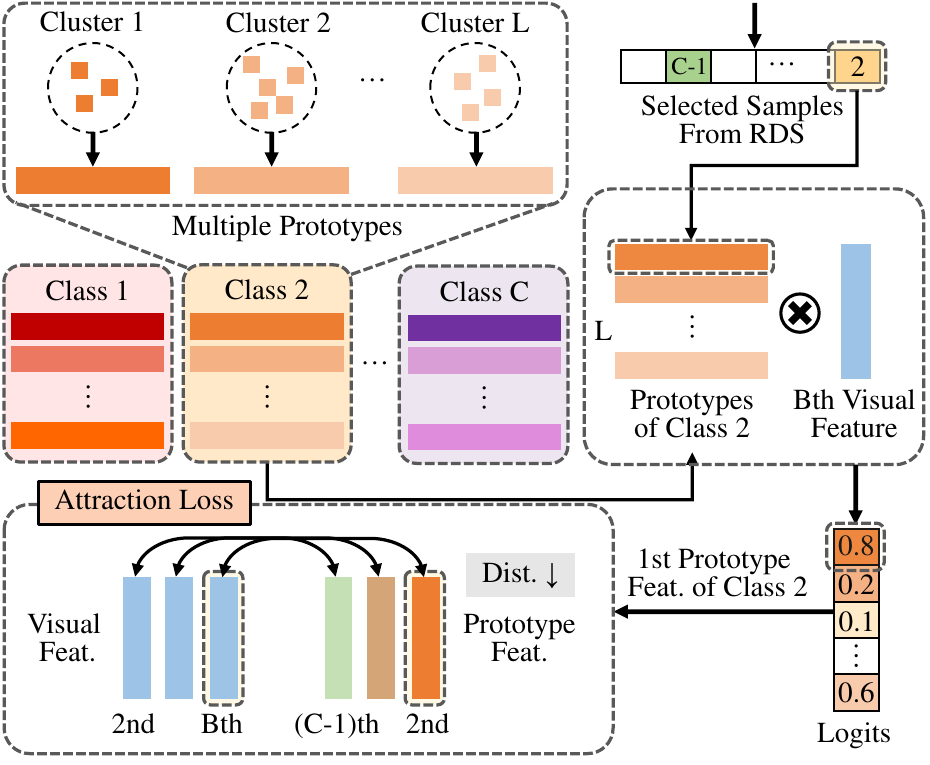} 
\caption{Proposed method employing unpaired images in the category-guided attraction loss. Unpaired images are further utilized to enhance object recognition performance and reconstructed image quality.
}
\vspace{-5pt}
\label{fig:prompt}
\end{figure}

\vspace{-2pt}
\label{sec:employ_unpair}
In this section, we expand the use case of our method by introducing how to employ the existing unpaired images.
Here, we assume that the category of each image is known, which is the common case for web crawling.
Although our framework is capable of event-to-image reconstruction and object recognition without using any images and labels, utilizing existing images can possibly improve the image reconstruction quality and consequently improve object recognition performance.
As shown in Fig.~\ref{fig:prompt}, we replace textual features in $\mathcal{L}_{att}$ with the prototype features generated from unpaired images.
For each category $c$, we first encode all images of category $c$ and apply a clustering algorithm to obtain $L$ clusters in total.
Then, we aggregate the features of each cluster by averaging them, resulting in $L$ prototype features $W^c = (w_i^c)_{i=1}^{L} \in \mathbb{R}^{L \times D}$.

For data sample index $i \in S_{\mathrm{RDS}}$ with corresponding visual feature $v_i$ and the predicted category $c_i$, we replace the textual features $f_{c_i}$ in Eq.~\ref{eq:mod_att_loss} to the closest prototype feature $w^{c_i}_{l_i}$, where the cluster index $l_i$ is calculated as,
\vspace{-2pt}
\begin{equation}
\label{eq:cluster_assign}
    l_i = \underset{j}{\mathrm{argmax}}(v_i (w_j^{c_i})^\top).
\vspace{-3pt}
\end{equation}
Our modification of $\mathcal{L}_{att}$ to employ unpaired images is then defined as follows:
\vspace{-2pt}
\begin{equation}
\label{eq:proto_att_loss}
    \mathcal{L}_{att} = -\sum_{i \in S_{\mathrm{RDS}}}{\log{\frac{\exp(v_i (w^{c_i}_{l_i})^\top)}{\sum_{j \in S_{\mathrm{RDS}}}{\exp(v_i (w^{c_j}_{l_j})^\top)}}}}, 
\vspace{-2pt}
\end{equation}

\vspace{-5pt}
\section{Datasets}
\vspace{-2pt}
\label{sec:dataset}
The experiments are conducted on two widely used event-based object recognition datasets, N-Caltech101~\cite{ncaltech} and N-ImageNet~\cite{kim2021n}, which are constructed by filming Caltech101~\cite{fei2004learning} and ImageNet~\cite{deng2009imagenet} datasets on LCD monitor with event camera.
Since N-Caltech101 and N-ImageNet datasets do not provide APS frames,
the image in the original dataset and corresponding event sequence have different resolutions and are not aligned in pixel-level. 
To create a visual prototype in Sec.~\ref{sec:employ_unpair}, we scrap images from the original dataset, mimicking the web crawling. 
To avoid including paired images of events in the train set, we randomly split the event train set in half for each class. 
One half is used to train the model, while the corresponding images from the other half are used to create the visual prototype.

\begin{table*}[t]
\begin{center}
\resizebox{.99\linewidth}{!}{
\begin{tabular}{lcccccccccc}
\toprule
& Image & Class & \multicolumn{3}{c}{N-Caltech101} & & \multicolumn{3}{c}{ N-ImageNet (Mini)}  \\
\cline { 4 - 6 } \cline { 8 - 10 }
Methods & Pair & Label & Accuracy $\uparrow$ & FID $\downarrow$ & IS $\uparrow$ & & Accuracy $\uparrow$ & FID $\downarrow$ & IS $\uparrow$ \\
\hline
\multicolumn{2}{l}{\textbf{Supervised Learning}} & & & & & \\
\hline
Histogram~\cite{maqueda2018event} & &\checkmark & \textit{63.8} & - & - && \textit{49.8} & - & -\\
Ev-gait~\cite{wang2019ev} & & \checkmark & \textit{61.9} & - & - && \textit{51.3} & - & -\\
EST~\cite{gehrig2019end} & &\checkmark & $\textit{81.7}^*$ & - & -  && \textit{52.5} & - & - \\
DiST~\cite{kim2021n} & &\checkmark & \textit{73.8} & - & -  && \textit{57.9} & - & - \\
\hline
\multicolumn{2}{l}{\textbf{Unsupervised Learning}} & & & & & & \\
\hline
HOTS~\cite{lagorce2016hots} & & & $\textit{21.0}^*$ & - & - && 1.6 & - & - \\
\hline 
\multicolumn{2}{l}{\textbf{Image Reconstruction + CLIP}} & & & & & & & & & \\
\hline
E2VID~\cite{e2vid} (Pre-train) & & & 64.0 & 83.90 & 12.81 && 13.3 & 147.14 & 7.82\\
E2VID~\cite{e2vid} (Fine-tune)& & & 59.8 & 85.31 & 12.91 && 13.4 & 144.57 & 7.48\\
E2VID~\cite{e2vid} (Fine-tune)& \checkmark & & 59.8 & 86.66 & 12.61 && 13.5 & \underline{144.03} & 7.40\\
E2VID~\cite{e2vid} (Fine-tune)& & \checkmark & 60.1 & 85.50 & 12.85 && 12.7 & 145.79 & 7.33\\
E2VID~\cite{e2vid} (Scratch)& \checkmark & & 9.4 & 302.48 & 2.60 && 1.0 & 299.70 & 1.15\\
SSL-E2VID~\cite{paredes2021back} (Pre-train) & & & 28.2	& 281.09 & 2.36 &&  1.6 & 324.43 & 2.08\\
SSL-E2VID~\cite{paredes2021back} (Scratch) & & & 30.5 & 228.30 &	4.37 && 1.7 & 250.55 & 3.00 \\
Wang~\etal~\cite{wang2019event} (Scratch) & & & 42.7 &	160.34 & 6.83 && 10.1 & 231.94 & 5.00 \\ 
Wang~\etal~\cite{wang2019event} (Scratch) & \checkmark && 43.5 & 72.67 & 4.50 && 19.0 & 180.76 & 4.89\\ 
Wang~\etal~\cite{wang2019event} (Scratch)& & \checkmark & 39.7 & 179.79 & 4.18 && 13.2 & 216.37 & 4.70\\
\hline
Ours (Text Prompt) & & & \underline{82.46} & \underline{62.29} & \underline{14.81} && 30.16 & 150.58 & \underline{7.92}\\
Ours (Visual Prototype) & & & \textbf{82.61} & \textbf{54.19} & \textbf{17.59}  & & \textbf{31.28} & \textbf{135.95} & \textbf{9.74}\\
\bottomrule
\end{tabular}
}
\end{center}
\vspace{-14pt}
\caption{The results of event-based object recognition and event-to-image reconstruction. The \textbf{bold} and \underline{underline} denote the best and the second-best performance except for supervised methods, respectively. ${}^{*}$ denotes the values taken from the original paper and \textit{italic} denotes the performance of supervised methods.
}
\label{tab:main_result}
\vspace{-10pt}
\end{table*}

\noindent
\textbf{N-Caltech101} dataset consists of 8,246 event stream data belonging to 101 different object classes. Each object is captured with $320\times245$ resolution (average) for 300 milliseconds. 
Existing works~\cite{wang2021joint, rebecq2019high} utilized this dataset to evaluate the performance of image reconstruction and event-based object recognition.

\noindent
\textbf{N-ImageNet} is recently released large-scale dataset, containing 1,781,167 event stream data of 1,000 object classes. Compared to N-Caltech101, it has a larger resolution of $640\times480$ and a very short capturing time with $5\times10^{-2}$ milliseconds.
Therefore, each event stream contains significantly sparse edge information.
N-ImageNet is too challenging for unsupervised studies without ground truth, so we utilize a randomly sampled subset of 100 classes called N-ImageNet (Mini). 
Details about the split of N-ImageNet (Mini) are provided in the \textit{supplementary material}.

\vspace{-2pt}
\section{Experiments}
\vspace{-2pt}

We compare our methods utilizing text prompts or
visual prototypes with other existing methods. 
We adopt four representations (Histogram~\cite{maqueda2018event}, Ev-gait~\cite{wang2019ev}, EST~\cite{gehrig2019end}, and DiST~\cite{kim2021n}) for supervised methods and an unsupervised method (HOTS~\cite{lagorce2016hots}).
Additionally, three image reconstruction networks, E2VID~\cite{e2vid}, SSL-E2VID~\cite{paredes2021back} and Wang~\etal~\cite{wang2019event}, are utilized with CLIP. 
For a fair comparison with image reconstruction networks, we record the performance of three different models for each image reconstruction network: the official pre-trained model, fine-tuned model, and a model trained from scratch. We evaluate these models under three conditions: paired images, images belonging to the same class, and randomly selected images.
Some cases are overlooked when the pre-trained model is unavailable or the training setup does not yield meaningful results.
For implementation details, see the \emph{supplementary material}.

\subsection{Label-free Object Recognition}
\vspace{-2pt}

\noindent
\textbf{Text Prompt.} The quantitative results are presented in Table~\ref{tab:main_result}.
The proposed method outperforms supervised object recognition methods on the N-Caltech101 dataset, despite not using category labels. Also, our method surpasses existing unsupervised methods and image reconstruction methods with CLIP by a large margin under all metrics.
Specifically, ours surpasses the best image reconstruction with CLIP method, E2VID (Pre-train), by around 28\% increase in accuracy even though it utilizes the image while pre-training.
On N-ImageNet, supervised methods record higher accuracy than ours since the dataset is challenging due to the sparse events captured within a short period. However, ours shows a significant improvement in all metrics compared to other unsupervised and image reconstruction-based methods.
Our approach demonstrates the ability to perform object recognition and image reconstruction from event data, even when trained solely on textual features without any image data.

\noindent
\textbf{Visual Prototype.} 
Besides text prompts, our framework can employ visual prototypes generated from scrapped images.
By incorporating visual information, our visual prototype-based approach exceeds the performance of relying solely on textual prompts and achieves the best performance among unsupervised and image reconstruction-based methods under all metrics across both datasets.
This demonstrates the superiority of the visual prototypes.

\begin{figure*}[t] 
\vspace{-2pt}
\centering 
\includegraphics[width=.97\linewidth]{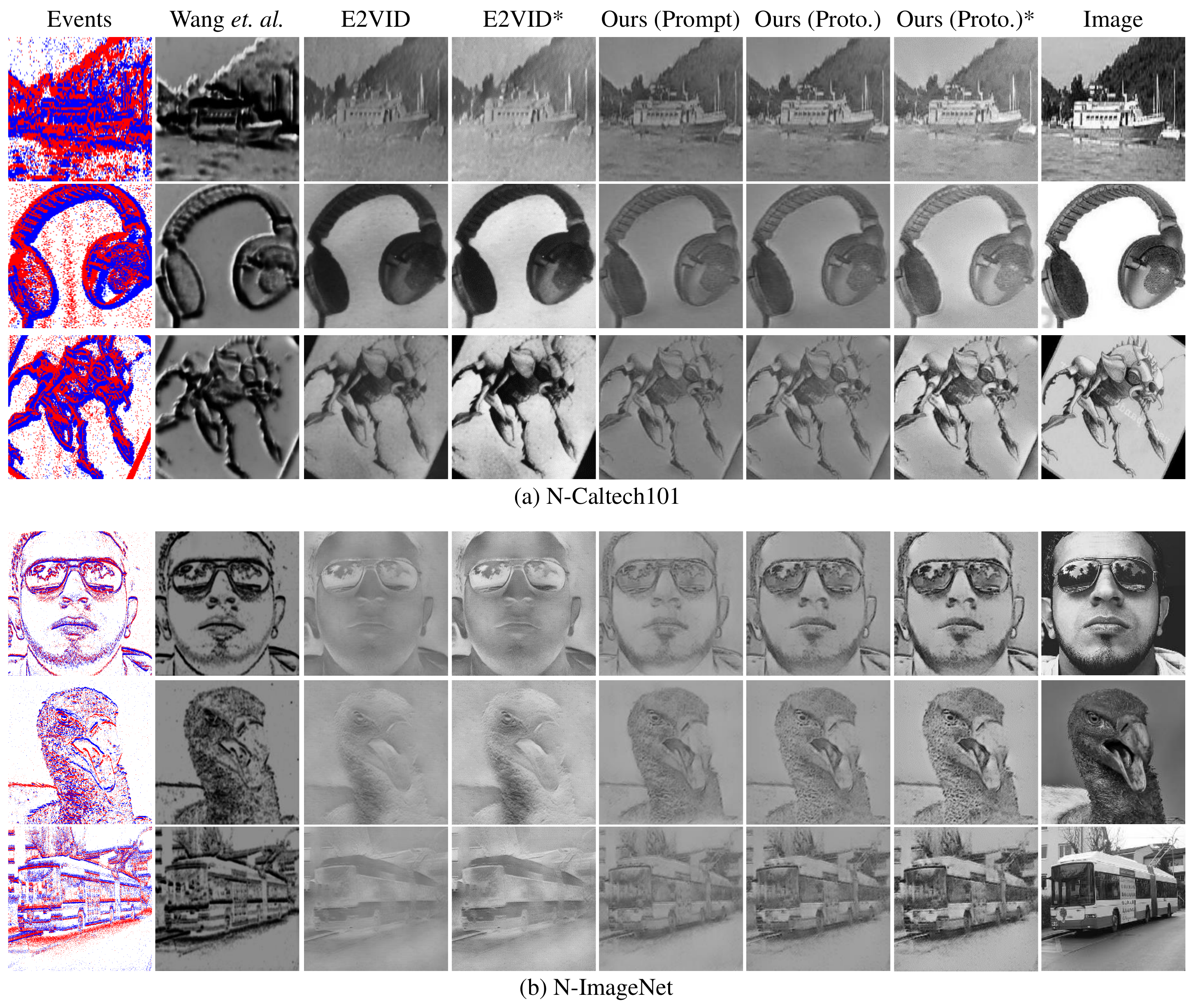} 

\vspace{-10pt}
\caption{Visual results of event-to-image reconstruction network on (a) N-Caltech101 and (b) N-ImageNet datasets. Ours shows a higher quality of reconstructed images with fine-grained details than other methods 
(Wang \etal \cite{wang2019event} and pre-trained E2VID \cite{e2vid}). * denotes the images with brightness and contrast adjustment.
}
\label{fig:e2i_qual}
\end{figure*}

\begin{table*}[t]
\begin{center}
\resizebox{.99\linewidth}{!}{
\begin{tabular}{lccccccccccc}
\toprule
& Image & Class & \multicolumn{3}{c}{N-Caltech101}  & & \multicolumn{3}{c}{ N-ImageNet (Mini)}  \\
\cline { 4 - 6 } \cline { 8 - 10 }
Methods & Pair & Label & Accuracy $\uparrow$ & FID $\downarrow$ & IS $\uparrow$ & & Accuracy $\uparrow$ & FID $\downarrow$ & IS $\uparrow$ \\
\hline 
\multicolumn{2}{l}{\textbf{Image Reconstruction + CLIP}} & & & & & & & & & \\
\hline
E2VID~\cite{e2vid} (Pre-train) & & & 58.4 & 194.26 & 6.47 && 31.7 & 203.05 & 6.64\\
E2VID~\cite{e2vid} (Fine-tune)& & & 61.7 & 203.39 & \underline{7.56} && 30.0 & 199.83 & 6.73 \\
E2VID~\cite{e2vid} (Fine-tune)& \checkmark & & 61.1 & 203.40 & 7.45 && 30.9	& \underline{199.44} & \underline{6.80}\\
E2VID~\cite{e2vid} (Fine-tune)& & \checkmark & 61.5 & 202.23 & \underline{7.56}&& 29.4	& 203.97 & 6.08\\
E2VID~\cite{e2vid} (Scratch)& \checkmark & & 21.3 &	362.08 & 2.43 && 4.9 & 333.13 & 1.36\\
SSL-E2VID~\cite{paredes2021back} (Pre-train) & & & 24.7	& 332.43 & 2.07 && 5.7 & 352.53 & 2.04 \\
SSL-E2VID~\cite{paredes2021back} (Scratch) & & & 25.3 & 289.36 & 3.93 && 6.1 & 287.98 & 2.79\\

Wang~\etal~\cite{wang2019event} (Scratch) & & & 47.5 & 228.77 & 4.67 && 18.7 & 241.87 & 6.12\\ 
Wang~\etal~\cite{wang2019event} (Scratch) & \checkmark && 48.5 & 230.74 & 4.91 && 20.5 & 236.78 & 3.90\\ 
Wang~\etal~\cite{wang2019event} (Scratch) & & \checkmark & 43.7 & 215.63 & 4.66 && 15.6& 237.38 & 4.40 \\
\hline
Ours (Text Prompt) & & & \underline{84.31} & \underline{147.31} & 6.81 && \underline{45.10} & 202.29 & 6.31\\
Ours (Visual Prototype) & & & \textbf{85.56} & \textbf{107.58} & \textbf{9.35} && \textbf{51.50} & \textbf{174.24} & \textbf{7.32}\\
\bottomrule
\end{tabular}
}
\end{center}
\vspace{-15pt}
\caption{The results of event-based zero-shot object recognition using CLIP.}
\vspace{-8pt}
\label{tab:zero_shot}
\end{table*}

\vspace{-2pt}
\subsection{Image-free Image Reconstruction}
\vspace{-2pt}
We also compare the reconstruction results of our methods with other event-to-image reconstruction networks in Table~\ref{tab:main_result}. Note that, as mentioned in Sec.~\ref{sec:dataset}, the resolution and pixel positions of events and corresponding original images are not aligned. 
Therefore, pixel-wise metrics (\eg,~L1, PSNR) cannot be used for comparison.
Instead, to assess the visual quality of the reconstructed images, we utilize two metrics: inception score (IS)~\cite{salimans2016improved} and Fréchet inception distance (FID)~\cite{heusel2017gans}.

Despite not utilizing any images during training, our text prompt-based approach surpasses all other image reconstruction techniques in terms of FID and IS metrics on the N-Caltech101 dataset. However, in the case of N-ImageNet, our text prompt approach is the strongest by an IS indicator but falls behind pre-trained E2VID in FID. Since ours is trained from events of short periods without paired images, it becomes challenging to learn the distribution of real images. Instead, utilizing the visual prototype results in the highest FID and IS indicators in both datasets. This implies that it can acquire the ability to reconstruct images without requiring paired images that correspond to events.


We also present the qualitative results in Fig.~\ref{fig:e2i_qual}. Wang~\etal~\cite{wang2019event} utilize the paired image in training but the method lacks the ability to effectively display intensity contrast, resulting in the loss of many details. While pre-trained E2VID shows better image quality over \cite{wang2019event}, it still struggles to preserve fine details as it operates in a distinct training environment. This is particularly apparent in the N-ImageNet dataset, where E2VID generates results that deviate significantly from the original image. Conversely, our framework is able to recover intensity effectively and preserve details in reconstructed images, even without access to paired images. Our results closely resemble real images, even in the challenging N-ImageNet dataset.

\subsection{Zero-shot Object Recognition}
\vspace{-2pt}
We further demonstrate the advantages of our framework over supervised methods by enabling it to be used for zero-shot object recognition. The reconstructed images can be utilized for zero-shot object recognition thanks to CLIP's capacity to align images with any semantic concepts in an open vocabulary for zero-shot object recognition. To define ``unseen" categories that are not seen during training, we use only 80 of the 100 categories for training and the remaining 20 for zero-shot testing.

In Table~\ref{tab:zero_shot}, we present the performance of zero-shot object recognition. Without any category labels, our methods with text prompt achieve a promising $84.31 \%$ and $45.10 \%$ on N-Caltech101 and N-ImageNet, respectively. This is 25.91$\%$ and 13.4$\%$ higher than E2VID pretrained with a large number of image pairs. 
These results validate that although our model learns through the process of being category-aware, generalization is possible for other categories, especially ``unseen" in this case. 
Although FID and IS of our method using text prompts are relatively poor compared to pre-trained E2VID on N-ImageNet, this can also be overcome by using visual prototypes. Our method using the visual prototype is able to achieve a substantial improvement under all metrics. Specifically, in the N-Caltech101 dataset, our method increases the IS from 6.81 to 9.35, while in the N-ImageNet dataset, it increases the accuracy from 45.10 to 51.50.

\begin{table}[t]
\begin{center}
\resizebox{.99\linewidth}{!}{
\begin{tabular}{lccccccc}
\toprule
& \multirow{2}{*}{$\mathcal{L}_{att}$}&\multirow{2}{*}{$\mathcal{L}_{rep}$} &\multirow{2}{*}{$\mathcal{L}_{con}$} & \multicolumn{2}{c}{RDS} & \multirow{2}{*}{Accuracy $\uparrow$}\\ 
\cline{5-6}& & &  & PPI & TRCI  &  \\
\hline 
(1) & \checkmark & & & & & 9.50\\
(2) & \checkmark & \checkmark & & & & 14.00 \\
\hline
(3) & \checkmark & \checkmark & \checkmark & & & 15.68 \\
(4) & \checkmark & \checkmark & & \checkmark & & 28.42 \\
(5) & \checkmark & \checkmark & \checkmark & \checkmark & & 28.90 \\
(6) & \checkmark & \checkmark & & \checkmark & \checkmark & 28.91\\
(7) & \checkmark & \checkmark & \checkmark & \checkmark & \checkmark & 30.16\\
\bottomrule
\end{tabular}
}
\end{center}
\vspace{-15pt}
\caption{Ablation study of the proposed method.}
\label{tab:able}
\end{table}

\section{Ablation Study and Dicussion}
\subsection{Ablation Study}
We conduct an ablation study to understand how the components influence performance. Table~\ref{tab:able} illustrates the effectiveness of each component on N-ImageNet dataset. We start with a baseline, which denotes the removal of everything except attraction loss, $\mathcal{L}_{att}$. During the experiment, we notice that if we only apply attraction loss towards the network without repulsion loss, $\mathcal{L}_{rep}$, the collapse often takes place. Therefore, $\mathcal{L}_{rep}$ is essential for stable learning of $\mathcal{L}_{att}$. It is worth mentioning the impact of the proposed PPI on performance. Upon analyzing (4), which displays the results when PPI is added to (2), it can be seen that the accuracy has increased by 14.42. Upon examining (5) and (7), it can be observed that the accuracy increases by 1.26, validating the impact of the sub-sampling process via TRCI. Furthermore, the ablation results between (6) and (7) show that the reconstruction consistency between global and local preserves the spatial details of the image, leading to performance improvement. Finally, the model (7) that has all of its components incorporated achieves the highest performance.

\vspace{-2pt}
\subsection{Number of Samples in PPI}
\begin{figure}[t] 
\centering 
\includegraphics[width=.99\linewidth]{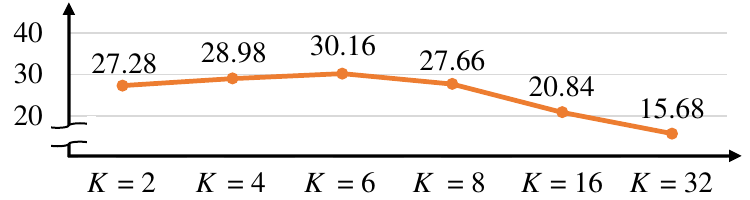} 
\caption{Object recognition accuracy on N-ImageNet (Mini) according to $K$ in PPI.}
\label{fig:k_mask}
\vspace{-4pt}
\end{figure}

\begin{figure}[t] 
\centering 
\includegraphics[width=1.0\linewidth]{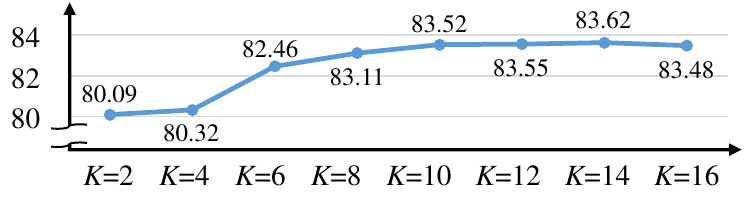} 
\vspace{-16pt}
\caption{Object recognition accuracy on N-Caltech101
according to $K$ in PPI.
}
\label{fig:k_mask_caltech}
\end{figure}

\vspace{-2pt}
The $K$ used in PPI for selecting the sample with the highest probability impacts the performance of object recognition. Fig.~\ref{fig:k_mask} shows the evaluation of performance on N-ImageNet dataset based on different values of $K$. When $K$ is set to 32, which includes all samples in the batch, the attraction loss, $\mathcal{L}_{att}$, is applied to even those samples with low reliability, resulting in the poorest performance. Due to the same reason, the performance is better when $K$ is 16 than when it is 32, but since unreliable samples are also selected, the accuracy is relatively lower than in other favorable situations. As $K$ decreases, the performance improves, but it starts to decline when $K$ is less than 6, indicating that even with high reliability, it is challenging to achieve optimal training if $\mathcal{L}_{att}$ is applied with very few samples. Similarly, we report the accuracy according to the $K$ value on the N-Caltech101 in Fig~\ref{fig:k_mask_caltech}.
Unless an extremely small number of samples are drawn, as the value of $K$ increases, the performance also improves. However, once a certain threshold is reached, there is little significant change, and it starts to decline with further increases. The accuracy exceeds 80\% in all cases, showing robustness to $K$ values.
Although the level of sensitivity varies depending on the dataset, we do confirm that satisfactory results can be obtained between $K=4$ and $K=10$.

\subsection{Hyper-parameter Analysis}
\begin{table}[t]
\label{tab:hyper}
\begin{center}
\resizebox{0.99\linewidth}{!}{\renewcommand{\tabcolsep}{1.6pt}
\begin{tabular}{c|ccc|ccc}
\hline
                     & \multicolumn{3}{c|}{N-Caltech101} & \multicolumn{3}{c}{N-ImageNet~(Mini)} \\ \hline
$\lambda_2$ \textbackslash{} $\lambda_3$ & 0.5      & 1       & 2       & 0.5      & 1       & 2       \\ \hline
0.001                & \gradient{82.71}    & \gradient{82.93}   & \gradient{81.26}   & \gradienttwo{29.62}    & \gradienttwo{26.34}   & \gradienttwo{30.08}   \\
0.005                & \gradient{83.52}    & \gradient{82.02}   & \gradient{81.66}   & \gradienttwo{29.38}    & \gradienttwo{26.92}   & \gradienttwo{28.72}   \\
0.01                 & \gradient{82.24}    & \gradient{82.46}   & \gradient{81.99}   & \gradienttwo{29.34}    & \gradienttwo{30.16}   & \gradienttwo{29.80}   \\
0.05                 & \gradient{81.99}    & \gradient{82.39}   & \gradient{82.61}   & \gradienttwo{27.50}    & \gradienttwo{27.92}   & \gradienttwo{29.00}   \\
0.1                  & \gradient{81.37}    & \gradient{82.46}   & \gradient{82.17}   & \gradienttwo{25.76}    & \gradienttwo{26.28}   & \gradienttwo{26.32}   \\ \hline
\end{tabular}}
\vspace{-5pt}
\caption{The accuracy according to hyper-paramter in Eq.~\ref{eq:total_loss} when $\lambda_1=1$.}
\label{tab:hyp}
\end{center}
\vspace{-12pt}
\end{table}

In Table~\ref{tab:hyp}, we report the accuracy according to $\lambda_2$ and $\lambda_3$ in Eq.~\ref{eq:total_loss}.
If the weight of the repulsion loss, $\lambda_2$, is set too high, the performance decreases. However, this can be mitigated by increasing the weight of the local-global reconstruction consistency loss, $\lambda_3$. The results verify the robustness of our method according to the $\lambda$ on both datasets.

\begin{figure}[t] 
\centering 
\includegraphics[width=.99\linewidth]{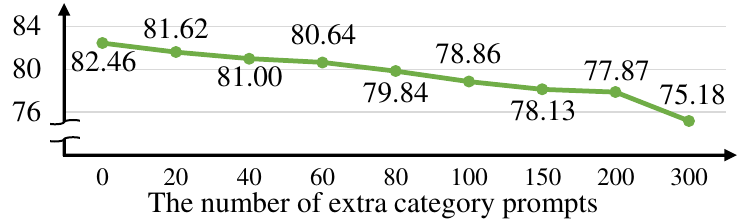} 
\vspace{-6pt}
\caption{Object recognition accuracy on N-Caltech101 according to the number of extra category prompts.}
\label{fig:superset}
\vspace{-4pt}
\end{figure}

\vspace{-2pt}
\subsection{Learning with Superset Categories}
\vspace{-2pt}
In practical terms, it can be challenging for users to find corresponding categories of the event dataset. As a result, they often end up taking more categories as prompts. We tackle these issues by demonstrating that our joint learning framework works robustly even when employing more prompts than the categories of the event on the N-Caltech101 dataset. To avoid using additional prompts that are too irrelevant to the N-Caltech101 dataset, we randomly take additional prompts from N-ImageNet. We show the performance results according to the number of extra category prompts in Fig.~\ref{fig:superset}. While the performance decreases with an increase in extra category prompts, it is worth noting that our algorithm remains reliable even when faced with over 100 additional prompts beyond the corresponding category. This suggests that our algorithm can function with superset categories, even in a situation when the number of extra categories is much larger than the actual categories.

\vspace{-2pt}
\section{Conclusion}
\vspace{-2pt}
In this work, we propose a joint learning framework for event-based object recognition with image reconstruction from events without labels and paired images.
With the proposed losses and the reliable data sampling strategy, our approach allows stable learning of two tasks simultaneously from pseudo labels.
As a result, our approach achieves exceptional performance in event-based object recognition and event-to-image reconstruction, all without requiring category labels and paired images.
Our approach successfully addresses the challenge of label dependency in event-based research, which has been a significant obstacle.
We anticipate that our work can be further extended to other tasks such as event-based zero-shot object detection, zero-shot semantic segmentation, etc., that have not explored so far in the research field for event cameras.

\noindent\textbf{Acknowledgements.} This work was supported by the National Research Foundation of Korea(NRF) grant funded by the Korea government(MSIT) (NRF-2022R1A2B5B03002636).

{\small
\bibliographystyle{ieee_fullname}
\bibliography{main}

\begin{thebibliography}{10}\itemsep=-1pt

\bibitem{amir2017low}
Arnon Amir, Brian Taba, David Berg, Timothy Melano, Jeffrey McKinstry, Carmelo
  Di~Nolfo, Tapan Nayak, Alexander Andreopoulos, Guillaume Garreau, Marcela
  Mendoza, et~al.
\newblock A low power, fully event-based gesture recognition system.
\newblock In {\em Proceedings of the IEEE conference on computer vision and
  pattern recognition}, pages 7243--7252, 2017.

\bibitem{bardow2016simultaneous}
Patrick Bardow, Andrew~J Davison, and Stefan Leutenegger.
\newblock Simultaneous optical flow and intensity estimation from an event
  camera.
\newblock In {\em Proceedings of the IEEE conference on computer vision and
  pattern recognition}, pages 884--892, 2016.

\bibitem{bi2019graph}
Yin Bi, Aaron Chadha, Alhabib Abbas, Eirina Bourtsoulatze, and Yiannis
  Andreopoulos.
\newblock Graph-based object classification for neuromorphic vision sensing.
\newblock In {\em Proceedings of the IEEE/CVF International Conference on
  Computer Vision}, pages 491--501, 2019.

\bibitem{cannici2020differentiable}
Marco Cannici, Marco Ciccone, Andrea Romanoni, and Matteo Matteucci.
\newblock A differentiable recurrent surface for asynchronous event-based data.
\newblock In {\em Computer Vision--ECCV 2020: 16th European Conference,
  Glasgow, UK, August 23--28, 2020, Proceedings, Part XX 16}, pages 136--152.
  Springer, 2020.

\bibitem{vit}
Nicolas Carion, Francisco Massa, Gabriel Synnaeve, Nicolas Usunier, Alexander
  Kirillov, and Sergey Zagoruyko.
\newblock End-to-end object detection with transformers.
\newblock In {\em Computer Vision--ECCV 2020: 16th European Conference,
  Glasgow, UK, August 23--28, 2020, Proceedings, Part I 16}, pages 213--229.
  Springer, 2020.

\bibitem{chefer2022image}
Hila Chefer, Sagie Benaim, Roni Paiss, and Lior Wolf.
\newblock Image-based clip-guided essence transfer.
\newblock In {\em Computer Vision--ECCV 2022: 17th European Conference, Tel
  Aviv, Israel, October 23--27, 2022, Proceedings, Part XIII}, pages 695--711.
  Springer, 2022.

\bibitem{chen2023clip2scene}
Runnan Chen, Youquan Liu, Lingdong Kong, Xinge Zhu, Yuexin Ma, Yikang Li,
  Yuenan Hou, Yu Qiao, and Wenping Wang.
\newblock Clip2scene: Towards label-efficient 3d scene understanding by clip.
\newblock {\em arXiv preprint arXiv:2301.04926}, 2023.

\bibitem{cheng2019det}
Wensheng Cheng, Hao Luo, Wen Yang, Lei Yu, Shoushun Chen, and Wei Li.
\newblock Det: A high-resolution dvs dataset for lane extraction.
\newblock In {\em Proceedings of the IEEE/CVF Conference on Computer Vision and
  Pattern Recognition Workshops}, pages 0--0, 2019.

\bibitem{cho2023learning}
Hoonhee Cho, Jegyeong Cho, and Kuk-Jin Yoon.
\newblock Learning adaptive dense event stereo from the image domain.
\newblock In {\em Proceedings of the IEEE/CVF Conference on Computer Vision and
  Pattern Recognition}, pages 17797--17807, 2023.

\bibitem{cho2021eomvs}
Hoonhee Cho, Jaeseok Jeong, and Kuk-Jin Yoon.
\newblock Eomvs: Event-based omnidirectional multi-view stereo.
\newblock {\em IEEE Robotics and Automation Letters}, 6(4):6709--6716, 2021.

\bibitem{cho2022event}
Hoonhee Cho and Kuk-Jin Yoon.
\newblock Event-image fusion stereo using cross-modality feature propagation.
\newblock In {\em Proceedings of the AAAI Conference on Artificial
  Intelligence}, volume~36, pages 454--462, 2022.

\bibitem{cho2022selection}
Hoonhee Cho and Kuk-Jin Yoon.
\newblock Selection and cross similarity for event-image deep stereo.
\newblock In {\em European Conference on Computer Vision}, pages 470--486.
  Springer, 2022.

\bibitem{cohen2018spatial}
Gregory Cohen, Saeed Afshar, Garrick Orchard, Jonathan Tapson, Ryad Benosman,
  and Andre van Schaik.
\newblock Spatial and temporal downsampling in event-based visual
  classification.
\newblock {\em IEEE Transactions on Neural Networks and Learning Systems},
  29(10):5030--5044, 2018.

\bibitem{cook2011interacting}
Matthew Cook, Luca Gugelmann, Florian Jug, Christoph Krautz, and Angelika
  Steger.
\newblock Interacting maps for fast visual interpretation.
\newblock In {\em The 2011 International Joint Conference on Neural Networks},
  pages 770--776. IEEE, 2011.

\bibitem{deng2009imagenet}
Jia Deng, Wei Dong, Richard Socher, Li-Jia Li, Kai Li, and Li Fei-Fei.
\newblock Imagenet: A large-scale hierarchical image database.
\newblock In {\em 2009 IEEE conference on computer vision and pattern
  recognition}, pages 248--255. Ieee, 2009.

\bibitem{fei2004learning}
Li Fei-Fei, Rob Fergus, and Pietro Perona.
\newblock Learning generative visual models from few training examples: An
  incremental bayesian approach tested on 101 object categories.
\newblock In {\em 2004 conference on computer vision and pattern recognition
  workshop}, pages 178--178. IEEE, 2004.

\bibitem{gehrig2019end}
Daniel Gehrig, Antonio Loquercio, Konstantinos~G Derpanis, and Davide
  Scaramuzza.
\newblock End-to-end learning of representations for asynchronous event-based
  data.
\newblock In {\em Proceedings of the IEEE/CVF International Conference on
  Computer Vision}, pages 5633--5643, 2019.

\bibitem{gehrig2018asynchronous}
Daniel Gehrig, Henri Rebecq, Guillermo Gallego, and Davide Scaramuzza.
\newblock Asynchronous, photometric feature tracking using events and frames.
\newblock In {\em Proceedings of the European Conference on Computer Vision
  (ECCV)}, pages 750--765, 2018.

\bibitem{he2016deep}
Kaiming He, Xiangyu Zhang, Shaoqing Ren, and Jian Sun.
\newblock Deep residual learning for image recognition.
\newblock In {\em Proceedings of the IEEE conference on computer vision and
  pattern recognition}, pages 770--778, 2016.

\bibitem{heusel2017gans}
Martin Heusel, Hubert Ramsauer, Thomas Unterthiner, Bernhard Nessler, and Sepp
  Hochreiter.
\newblock Gans trained by a two time-scale update rule converge to a local nash
  equilibrium.
\newblock {\em Advances in neural information processing systems}, 30, 2017.

\bibitem{hu2016dvs}
Yuhuang Hu, Hongjie Liu, Michael Pfeiffer, and Tobi Delbruck.
\newblock Dvs benchmark datasets for object tracking, action recognition, and
  object recognition.
\newblock {\em Frontiers in neuroscience}, 10:405, 2016.

\bibitem{kim2008simultaneous}
Hanme Kim, Ankur Handa, Ryad Benosman, Sio-Hoi Ieng, and Andrew~J Davison.
\newblock Simultaneous mosaicing and tracking with an event camera.
\newblock {\em J. Solid State Circ}, 43:566--576, 2008.

\bibitem{kim2016real}
Hanme Kim, Stefan Leutenegger, and Andrew~J Davison.
\newblock Real-time 3d reconstruction and 6-dof tracking with an event camera.
\newblock In {\em Computer Vision--ECCV 2016: 14th European Conference,
  Amsterdam, The Netherlands, October 11-14, 2016, Proceedings, Part VI 14},
  pages 349--364. Springer, 2016.

\bibitem{kim2021n}
Junho Kim, Jaehyeok Bae, Gangin Park, Dongsu Zhang, and Young~Min Kim.
\newblock N-imagenet: Towards robust, fine-grained object recognition with
  event cameras.
\newblock In {\em Proceedings of the IEEE/CVF International Conference on
  Computer Vision}, pages 2146--2156, 2021.

\bibitem{kim2023event}
Taewoo Kim, Yujeong Chae, Hyun-Kurl Jang, and Kuk-Jin Yoon.
\newblock Event-based video frame interpolation with cross-modal asymmetric
  bidirectional motion fields.
\newblock In {\em Proceedings of the IEEE/CVF Conference on Computer Vision and
  Pattern Recognition}, pages 18032--18042, 2023.

\bibitem{klenk2022masked}
Simon Klenk, David Bonello, Lukas Koestler, and Daniel Cremers.
\newblock Masked event modeling: Self-supervised pretraining for event cameras.
\newblock {\em arXiv preprint arXiv:2212.10368}, 2022.

\bibitem{lagorce2016hots}
Xavier Lagorce, Garrick Orchard, Francesco Galluppi, Bertram~E Shi, and Ryad~B
  Benosman.
\newblock Hots: a hierarchy of event-based time-surfaces for pattern
  recognition.
\newblock {\em IEEE transactions on pattern analysis and machine intelligence},
  39(7):1346--1359, 2016.

\bibitem{lee2016training}
Jun~Haeng Lee, Tobi Delbruck, and Michael Pfeiffer.
\newblock Training deep spiking neural networks using backpropagation.
\newblock {\em Frontiers in neuroscience}, 10:508, 2016.

\bibitem{li2017cifar10}
Hongmin Li, Hanchao Liu, Xiangyang Ji, Guoqi Li, and Luping Shi.
\newblock Cifar10-dvs: an event-stream dataset for object classification.
\newblock {\em Frontiers in neuroscience}, 11:309, 2017.

\bibitem{lungu2017live}
Iulia-Alexandra Lungu, Federico Corradi, and Tobi Delbr{\"u}ck.
\newblock Live demonstration: Convolutional neural network driven by dynamic
  vision sensor playing roshambo.
\newblock In {\em 2017 IEEE International Symposium on Circuits and Systems
  (ISCAS)}, pages 1--1. IEEE, 2017.

\bibitem{maqueda2018event}
Ana~I Maqueda, Antonio Loquercio, Guillermo Gallego, Narciso Garc{\'\i}a, and
  Davide Scaramuzza.
\newblock Event-based vision meets deep learning on steering prediction for
  self-driving cars.
\newblock In {\em Proceedings of the IEEE conference on computer vision and
  pattern recognition}, pages 5419--5427, 2018.

\bibitem{messikommer2020event}
Nico Messikommer, Daniel Gehrig, Antonio Loquercio, and Davide Scaramuzza.
\newblock Event-based asynchronous sparse convolutional networks.
\newblock In {\em Computer Vision--ECCV 2020: 16th European Conference,
  Glasgow, UK, August 23--28, 2020, Proceedings, Part VIII 16}, pages 415--431.
  Springer, 2020.

\bibitem{moeys2018pred18}
Diederik~Paul Moeys, Daniel Neil, Federico Corradi, Emmett Kerr, Philip Vance,
  Gautham Das, Sonya~A Coleman, Thomas~M McGinnity, Dermot Kerr, and Tobi
  Delbruck.
\newblock Pred18: Dataset and further experiments with davis event camera in
  predator-prey robot chasing.
\newblock {\em arXiv preprint arXiv:1807.03128}, 2018.

\bibitem{neil2016phased}
Daniel Neil, Michael Pfeiffer, and Shih-Chii Liu.
\newblock Phased lstm: Accelerating recurrent network training for long or
  event-based sequences.
\newblock {\em Advances in neural information processing systems}, 29, 2016.

\bibitem{infonce}
Aaron van~den Oord, Yazhe Li, and Oriol Vinyals.
\newblock Representation learning with contrastive predictive coding.
\newblock {\em arXiv preprint arXiv:1807.03748}, 2018.

\bibitem{orchard2013spiking}
Garrick Orchard, Ryad Benosman, Ralph Etienne-Cummings, and Nitish~V Thakor.
\newblock A spiking neural network architecture for visual motion estimation.
\newblock In {\em 2013 IEEE Biomedical Circuits and Systems Conference
  (BioCAS)}, pages 298--301. IEEE, 2013.

\bibitem{ncaltech}
Garrick Orchard, Ajinkya Jayawant, Gregory~K Cohen, and Nitish Thakor.
\newblock Converting static image datasets to spiking neuromorphic datasets
  using saccades.
\newblock {\em Frontiers in neuroscience}, 9:437, 2015.

\bibitem{orchard2015hfirst}
Garrick Orchard, Cedric Meyer, Ralph Etienne-Cummings, Christoph Posch, Nitish
  Thakor, and Ryad Benosman.
\newblock Hfirst: A temporal approach to object recognition.
\newblock {\em IEEE transactions on pattern analysis and machine intelligence},
  37(10):2028--2040, 2015.

\bibitem{paredes2021back}
Federico Paredes-Vall{\'e}s and Guido~CHE de Croon.
\newblock Back to event basics: Self-supervised learning of image
  reconstruction for event cameras via photometric constancy.
\newblock In {\em Proceedings of the IEEE/CVF Conference on Computer Vision and
  Pattern Recognition}, pages 3446--3455, 2021.

\bibitem{park2016performance}
Paul~KJ Park, Baek~Hwan Cho, Jin~Man Park, Kyoobin Lee, Ha~Young Kim, Hyo~Ah
  Kang, Hyun~Goo Lee, Jooyeon Woo, Yohan Roh, Won~Jo Lee, et~al.
\newblock Performance improvement of deep learning based gesture recognition
  using spatiotemporal demosaicing technique.
\newblock In {\em 2016 IEEE International Conference on Image Processing
  (ICIP)}, pages 1624--1628. IEEE, 2016.

\bibitem{paszke2019pytorch}
Adam Paszke, Sam Gross, Francisco Massa, Adam Lerer, James Bradbury, Gregory
  Chanan, Trevor Killeen, Zeming Lin, Natalia Gimelshein, Luca Antiga, et~al.
\newblock Pytorch: An imperative style, high-performance deep learning library.
\newblock {\em Advances in neural information processing systems}, 32, 2019.

\bibitem{patashnik2021styleclip}
Or Patashnik, Zongze Wu, Eli Shechtman, Daniel Cohen-Or, and Dani Lischinski.
\newblock Styleclip: Text-driven manipulation of stylegan imagery.
\newblock In {\em Proceedings of the IEEE/CVF International Conference on
  Computer Vision}, pages 2085--2094, 2021.

\bibitem{perez2013mapping}
Jos{\'e}~Antonio P{\'e}rez-Carrasco, Bo Zhao, Carmen Serrano, Begona Acha,
  Teresa Serrano-Gotarredona, Shouchun Chen, and Bernab{\'e} Linares-Barranco.
\newblock Mapping from frame-driven to frame-free event-driven vision systems
  by low-rate rate coding and coincidence processing--application to
  feedforward convnets.
\newblock {\em IEEE transactions on pattern analysis and machine intelligence},
  35(11):2706--2719, 2013.

\bibitem{pini2018learn}
Stefano Pini, Guido Borghi, and Roberto Vezzani.
\newblock Learn to see by events: Color frame synthesis from event and rgb
  cameras.
\newblock {\em arXiv preprint arXiv:1812.02041}, 2018.

\bibitem{radford2021learning_clip}
Alec Radford, Jong~Wook Kim, Chris Hallacy, Aditya Ramesh, Gabriel Goh,
  Sandhini Agarwal, Girish Sastry, Amanda Askell, Pamela Mishkin, Jack Clark,
  et~al.
\newblock Learning transferable visual models from natural language
  supervision.
\newblock In {\em International conference on machine learning}, pages
  8748--8763. PMLR, 2021.

\bibitem{ramesh2020low}
Bharath Ramesh, Andr{\'e}s Ussa, Luca Della~Vedova, Hong Yang, and Garrick
  Orchard.
\newblock Low-power dynamic object detection and classification with freely
  moving event cameras.
\newblock {\em Frontiers in neuroscience}, 14:135, 2020.

\bibitem{e2vid}
Henri Rebecq, Ren{\'e} Ranftl, Vladlen Koltun, and Davide Scaramuzza.
\newblock Events-to-video: Bringing modern computer vision to event cameras.
\newblock In {\em Proceedings of the IEEE/CVF Conference on Computer Vision and
  Pattern Recognition}, pages 3857--3866, 2019.

\bibitem{rebecq2019high}
Henri Rebecq, Ren{\'e} Ranftl, Vladlen Koltun, and Davide Scaramuzza.
\newblock High speed and high dynamic range video with an event camera.
\newblock {\em IEEE transactions on pattern analysis and machine intelligence},
  43(6):1964--1980, 2019.

\bibitem{reinbacher2016real}
Christian Reinbacher, Gottfried Graber, and Thomas Pock.
\newblock Real-time intensity-image reconstruction for event cameras using
  manifold regularisation.
\newblock {\em arXiv preprint arXiv:1607.06283}, 2016.

\bibitem{ronneberger2015u}
Olaf Ronneberger, Philipp Fischer, and Thomas Brox.
\newblock U-net: Convolutional networks for biomedical image segmentation.
\newblock In {\em Medical Image Computing and Computer-Assisted
  Intervention--MICCAI 2015: 18th International Conference, Munich, Germany,
  October 5-9, 2015, Proceedings, Part III 18}, pages 234--241. Springer, 2015.

\bibitem{salimans2016improved}
Tim Salimans, Ian Goodfellow, Wojciech Zaremba, Vicki Cheung, Alec Radford, and
  Xi Chen.
\newblock Improved techniques for training gans.
\newblock {\em Advances in neural information processing systems}, 29, 2016.

\bibitem{sanghi2022clip}
Aditya Sanghi, Hang Chu, Joseph~G Lambourne, Ye Wang, Chin-Yi Cheng, Marco
  Fumero, and Kamal~Rahimi Malekshan.
\newblock Clip-forge: Towards zero-shot text-to-shape generation.
\newblock In {\em Proceedings of the IEEE/CVF Conference on Computer Vision and
  Pattern Recognition}, pages 18603--18613, 2022.

\bibitem{scheerlinck2019continuous}
Cedric Scheerlinck, Nick Barnes, and Robert Mahony.
\newblock Continuous-time intensity estimation using event cameras.
\newblock In {\em Computer Vision--ACCV 2018: 14th Asian Conference on Computer
  Vision, Perth, Australia, December 2--6, 2018, Revised Selected Papers, Part
  V}, pages 308--324. Springer, 2019.

\bibitem{scheerlinck2020fast}
Cedric Scheerlinck, Henri Rebecq, Daniel Gehrig, Nick Barnes, Robert Mahony,
  and Davide Scaramuzza.
\newblock Fast image reconstruction with an event camera.
\newblock In {\em Proceedings of the IEEE/CVF Winter Conference on Applications
  of Computer Vision}, pages 156--163, 2020.

\bibitem{serrano2015poker}
Teresa Serrano-Gotarredona and Bernab{\'e} Linares-Barranco.
\newblock Poker-dvs and mnist-dvs. their history, how they were made, and other
  details.
\newblock {\em Frontiers in neuroscience}, 9:481, 2015.

\bibitem{shiba2022secrets}
Shintaro Shiba, Yoshimitsu Aoki, and Guillermo Gallego.
\newblock Secrets of event-based optical flow.
\newblock In {\em European Conference on Computer Vision}, pages 628--645.
  Springer, 2022.

\bibitem{sironi2018hats}
Amos Sironi, Manuele Brambilla, Nicolas Bourdis, Xavier Lagorce, and Ryad
  Benosman.
\newblock Hats: Histograms of averaged time surfaces for robust event-based
  object classification.
\newblock In {\em Proceedings of the IEEE conference on computer vision and
  pattern recognition}, pages 1731--1740, 2018.

\bibitem{sun2022event}
Lei Sun, Christos Sakaridis, Jingyun Liang, Qi Jiang, Kailun Yang, Peng Sun,
  Yaozu Ye, Kaiwei Wang, and Luc~Van Gool.
\newblock Event-based fusion for motion deblurring with cross-modal attention.
\newblock In {\em European Conference on Computer Vision}, pages 412--428.
  Springer, 2022.

\bibitem{tewel2022zerocap}
Yoad Tewel, Yoav Shalev, Idan Schwartz, and Lior Wolf.
\newblock Zerocap: Zero-shot image-to-text generation for visual-semantic
  arithmetic.
\newblock In {\em Proceedings of the IEEE/CVF Conference on Computer Vision and
  Pattern Recognition}, pages 17918--17928, 2022.

\bibitem{vasudevan2020introduction}
Ajay Vasudevan, Pablo Negri, Bernabe Linares-Barranco, and Teresa
  Serrano-Gotarredona.
\newblock Introduction and analysis of an event-based sign language dataset.
\newblock In {\em 2020 15th IEEE International Conference on Automatic Face and
  Gesture Recognition (FG 2020)}, pages 675--682. IEEE, 2020.

\bibitem{vaswani2017attention}
Ashish Vaswani, Noam Shazeer, Niki Parmar, Jakob Uszkoreit, Llion Jones,
  Aidan~N Gomez, {\L}ukasz Kaiser, and Illia Polosukhin.
\newblock Attention is all you need.
\newblock {\em Advances in neural information processing systems}, 30, 2017.

\bibitem{wang2021dual}
Lin Wang, Yujeong Chae, and Kuk-Jin Yoon.
\newblock Dual transfer learning for event-based end-task prediction via
  pluggable event to image translation.
\newblock In {\em Proceedings of the IEEE/CVF International Conference on
  Computer Vision}, pages 2135--2145, 2021.

\bibitem{wang2021evdistill}
Lin Wang, Yujeong Chae, Sung-Hoon Yoon, Tae-Kyun Kim, and Kuk-Jin Yoon.
\newblock Evdistill: Asynchronous events to end-task learning via bidirectional
  reconstruction-guided cross-modal knowledge distillation.
\newblock In {\em Proceedings of the IEEE/CVF Conference on Computer Vision and
  Pattern Recognition}, pages 608--619, 2021.

\bibitem{wang2019event}
Lin Wang, Yo-Sung Ho, Kuk-Jin Yoon, et~al.
\newblock Event-based high dynamic range image and very high frame rate video
  generation using conditional generative adversarial networks.
\newblock In {\em Proceedings of the IEEE/CVF Conference on Computer Vision and
  Pattern Recognition}, pages 10081--10090, 2019.

\bibitem{wang2021joint}
Lin Wang, Tae-Kyun Kim, and Kuk-Jin Yoon.
\newblock Joint framework for single image reconstruction and super-resolution
  with an event camera.
\newblock {\em IEEE Transactions on Pattern Analysis and Machine Intelligence},
  44(11):7657--7673, 2021.

\bibitem{wang2019ev}
Yanxiang Wang, Bowen Du, Yiran Shen, Kai Wu, Guangrong Zhao, Jianguo Sun, and
  Hongkai Wen.
\newblock Ev-gait: Event-based robust gait recognition using dynamic vision
  sensors.
\newblock In {\em Proceedings of the IEEE/CVF Conference on Computer Vision and
  Pattern Recognition}, pages 6358--6367, 2019.

\bibitem{weng2021event}
Wenming Weng, Yueyi Zhang, and Zhiwei Xiong.
\newblock Event-based video reconstruction using transformer.
\newblock In {\em Proceedings of the IEEE/CVF International Conference on
  Computer Vision}, pages 2563--2572, 2021.

\bibitem{lamb}
Yang You, Jing Li, Sashank Reddi, Jonathan Hseu, Sanjiv Kumar, Srinadh
  Bhojanapalli, Xiaodan Song, James Demmel, Kurt Keutzer, and Cho-Jui Hsieh.
\newblock Large batch optimization for deep learning: Training bert in 76
  minutes.
\newblock {\em arXiv preprint arXiv:1904.00962}, 2019.

\bibitem{zhu2019unsupervised}
Alex~Zihao Zhu, Liangzhe Yuan, Kenneth Chaney, and Kostas Daniilidis.
\newblock Unsupervised event-based learning of optical flow, depth, and
  egomotion.
\newblock In {\em Proceedings of the IEEE/CVF Conference on Computer Vision and
  Pattern Recognition}, pages 989--997, 2019.

\end{thebibliography}
}

\newpage







\maketitle
\def\thefootnote{*}\footnotetext{The first two authors contributed equally. In alphabetical order. }\def\thefootnote{\arabic{footnote}}
\ificcvfinal\thispagestyle{empty}\fi

This supplementary material provides how to employ web-crawled images, implementation and dataset details, and more qualitative results. In particular, the following contents are included in the supplementary material:

\begin{itemize}
\vspace{-3pt}
    \item Results of employing web-crawled images.
    \vspace{-3pt}
    \item Implementation details of the proposed method.
    \vspace{-3pt}
    \item Category split of datasets used in our experiments.
    \vspace{-3pt}
    \item Additional experiments and qualitative results.
\end{itemize}

\section{Describing Notations and Abbreviations}
We provide the descriptions of notations and abbreviations in Table~\ref{tab:abbrev} for better understanding.

\begin{table}[h]
\vspace{1pt}
\begin{center}
\resizebox{0.99\linewidth}{!}{\renewcommand{\tabcolsep}{6pt}
\begin{tabular}{cccc}
\hline
Abbrev. & Description & Abbrev. & Description \\ 
\hline
$\mathcal{I}$ & Intensity image & $\mathcal{L}_{att}$ & Category-guided attraction loss \\
$\mathcal{E}$ & Event stream & $\mathcal{L}_{rep}$ & Category-agnostic repulsion loss \\
$\mathcal{G}$ & Recon. network & PPI & Posterior probability indicator\\
$f_i$ & i-th textual feature & TRCI & Temporally reversed consistency indicator\\
$v_i$ & i-th visual feature & $S_{\mathrm{PPI}}$ & Selected sample indices from PPI\\
$p_i$ & i-th category prob. & $S_{\mathrm{TRCI}}$ & Selected sample indices from TRCI \\
$c$ & Predicted category & $S_{\mathrm{RDS}}$ & Intersection of $S_{\mathrm{PPI}}$ and $S_{\mathrm{TRCI}}$ \\
$p_c$ & Posterior prob. for c & $\mathcal{L}_{con}$ & Local-global recon. consistency loss  \\
$\mathcal{E}^R$ & Reversed event stream & $\mathcal{K}$ & Num. of data samples \\
$w_i$ & i-th prototype feature & $\mathcal{L}$ & Num. of clusters \\
\hline
\end{tabular}}
\end{center}
\vspace{-4pt}
\caption{Notions and abbreviations.}
\label{tab:abbrev}
\end{table}

\section{Employing Web-crawled Images}

\begin{figure}[h!] 
\centering 
\includegraphics[width=.99\columnwidth]{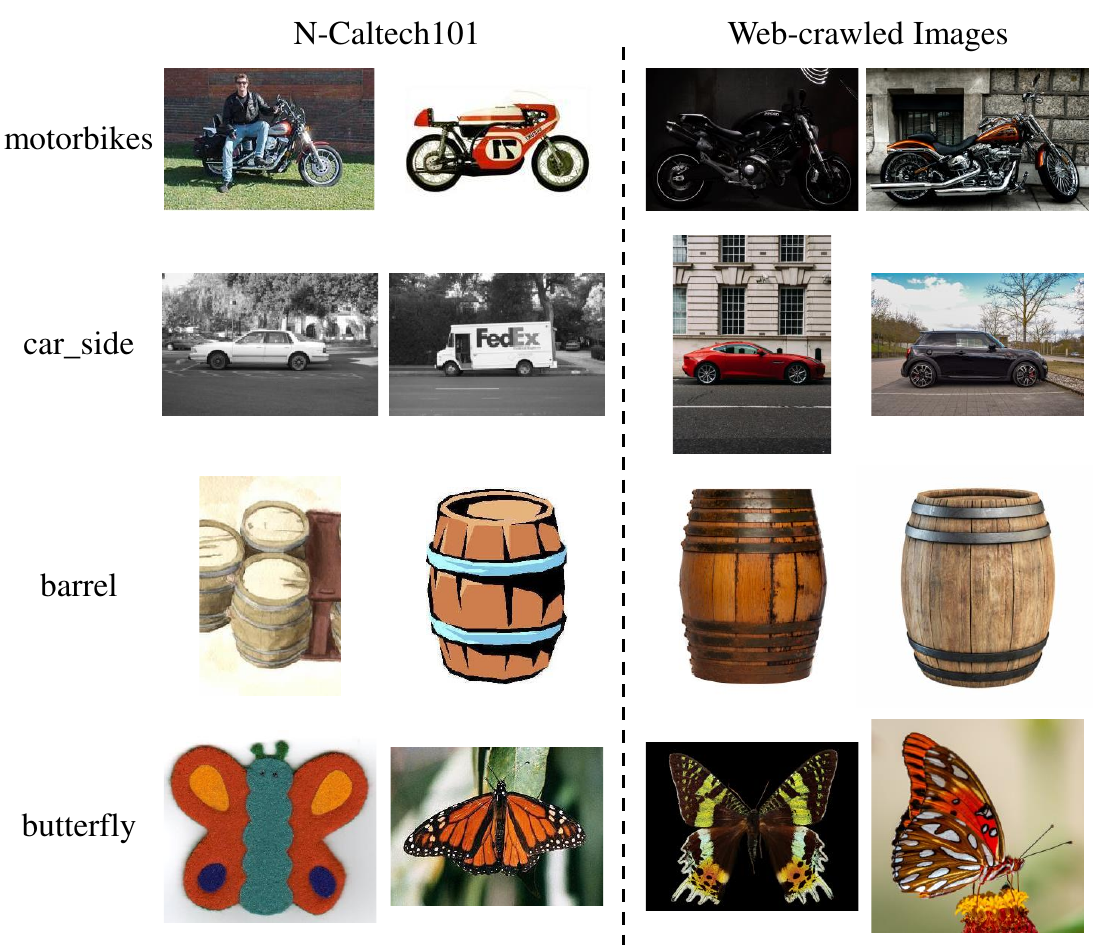} 
\caption{Examples from N-Caltech101 and web-crawled images of the corresponding categories. 
}
\label{fig:crawl}
\end{figure}

In the main paper, we split the dataset in half and used half of them as unpaired images of the other half, paying attention to the ease of experimentation or follow-up research.
In this section, instead of using the original dataset, we collected unpaired images by performing web crawling from google.
Examples of web-crawled images are shown in Fig.~\ref{fig:crawl}.
As shown in Table~\ref{tab:crawl_result}, we report the results and confirm that using web-crawled images is comparable and even better than using unpaired images from the original dataset in all metrics.
These findings affirm the practical efficacy of our approach in harnessing web crawling for improved performance.

\begin{table}[t]
\begin{center}
\resizebox{.99\linewidth}{!}{
\begin{tabular}{lccc}
\toprule
& \multicolumn{3}{c}{N-Caltech101}  \\
\cline { 2 - 4 }
Methods & Accuracy $\uparrow$ & FID $\downarrow$ & IS $\uparrow$ \\
\hline
Ours (Text Prompt) & 82.46 & 62.29 & 14.81 \\
Ours (Visual Prototype: N-Caltech101) & \underline{82.61} & \underline{54.19} & \underline{17.59}  \\
Ours (Visual Prototype: Web Crawling) & \textbf{82.64} & \textbf{51.83} & \textbf{18.25}  \\
\bottomrule
\end{tabular}
}
\end{center}
\vspace{-5pt}
\caption{The results of our method with web crawling.
}
\label{tab:crawl_result}
\end{table}

\begin{table*}[t]
\begin{center}
\resizebox{0.99\linewidth}{!}{\renewcommand{\tabcolsep}{6pt}
\begin{tabular}{c||c|c|c|c}
\hline
Dataset   & \multicolumn{2}{c|}{N-Caltech101} & \multicolumn{2}{c|}{N-ImageNet (Mini)} \\ \hline\hline
Setting  & standard & zero-shot & standard & zero-shot \\
\hline\hline
Train &  
\begin{tabular}[c]{@{}c@{}}
Faces\_easy, Leopards, Motorbikes, accordion, \\
airplanes, anchor, ant, barrel, \\
bass, beaver, binocular, bonsai,\\
brain, brontosaurus, buddha, butterfly,\\
camera, cannon, car\_side, ceiling\_fan, \\
cellphone, chair, chandelier, cougar\_body, \\
cougar\_face, crab, crayfish, crocodile,\\
crocodile\_head, cup, dalmatian, dollar\_bill,\\
dolphin, dragonfly, electric\_guitar, elephant,\\
emu, euphonium, ewer, ferry, \\
flamingo, flamingo\_head, garfield, gerenuk, \\
gramophone, grand\_piano, hawksbill, headphone, \\
hedgehog, helicopter, ibis, inline\_skate, \\
joshua\_tree, kangaroo, ketch, lamp, \\
laptop, llama, lobster, lotus, \\
mandolin, mayfly, menorah, metronome, \\
minaret, nautilus, octopus, okapi, \\
pagoda, panda, pigeon, pizza, \\
platypus, pyramid, revolver, rhino,\\
rooster, saxophone, schooner, scissors, \\
scorpion, sea\_horse, snoopy, soccer\_ball, \\
stapler, starfish, stegosaurus, stop\_sign, \\
strawberry, sunflower, tick, trilobite, \\
umbrella, watch, water\_lilly, wheelchair, \\
wild\_cat, windsor\_chair, wrench, yin\_yang \\
\end{tabular} 
& 
\begin{tabular}[c]{@{}c@{}}
Faces\_easy, Motorbikes, accordion, airplanes, \\
anchor, ant, barrel,  beaver, \\
binocular, bonsai, brain, brontosaurus, \\
buddha, butterfly, camera, cannon, \\
car\_side, ceiling\_fan, cellphone, chair, \\
chandelier, cougar\_body, cougar\_face, crab,\\
crayfish, crocodile\_head, cup, dalmatian,\\
dollar\_bill, dolphin,  electric\_guitar, elephant, \\
emu,  ewer, ferry,  flamingo\_head, \\
garfield, gerenuk, gramophone, grand\_piano, \\
hawksbill, helicopter,  inline\_skate, lobster, \\
lotus, mandolin, mayfly, menorah, \\
metronome,  nautilus, octopus, okapi,\\
pagoda, panda, pigeon, pizza, \\
platypus, pyramid, revolver, rhino, \\
rooster, schooner, scissors, scorpion, \\
snoopy, soccer\_ball, stapler, starfish, \\
stegosaurus, stop\_sign, strawberry, sunflower, \\
tick, trilobite, umbrella, watch, \\
water\_lilly, wheelchair, wild\_cat, windsor\_chair
\end{tabular} 
& 
\begin{tabular}[c]{@{}c@{}}
hamster, academic gown, airship, jackfruit, \\
barbershop,cocktail shaker, Komodo dragon, sunglasses,\\
grey fox,cello, comic book, goldfish, \\
Bloodhound, porcupine,jaguar, kingsnake, \\
altar, water buffalo, chiton, scarf,\\
storage chest, tool kit, sea anemone, \\
Border Terrier,menu, picket fence, forklift,\\
yellow ladys slipper, chameleon, dragonfly, Pomeranian, \\
European garden spider,Airedale Terrier, \\
frilled-necked lizard, black stork,\\
valley, radio telescope, leopard, crossword,\\
Australian Terrier, Shih Tzu, husky, can opener,\\
artichoke, assault rifle, fountain pen, harvestman,\\
parallel bars, harmonica, half-track, snoek fish,\\
pencil sharpener, submarine, muzzle, \\
eastern diamondback rattlesnake, \\
Miniature Schnauzer,missile, Komondor, grand piano,\\
website, king penguin,canoe, red-breasted merganser, \\
trolleybus, quail,poke bonnet, King Charles Spaniel, race car, \\
Malinois, solar thermal collector, slug, bucket, \\
dung beetle,Asian elephant, window screen, Flat-Coated Retriever,\\
steel drum, snowplow, handkerchief, tailed frog, \\
church,Chesapeake Bay Retriever, Christmas stocking, hatchet,\\
hair clip, vulture, sidewinder rattlesnake, oscilloscope, \\
worm snake, eel, wok, planetarium, \\
Old English Sheepdog,platypus, Pembroke Welsh Corgi,\\
alligator lizard, consomme, African rock python, \\
hot tub, Tibetan Mastiff\\
\end{tabular}
& 
\begin{tabular}[c]{@{}c@{}}
hamster, academic gown, jackfruit, barbershop, \\
 Komodo dragon, sunglasses, grey fox, cello, \\
 comic book, goldfish, Bloodhound, porcupine, \\
 jaguar, altar, water buffalo, chiton, \\
 scarf, storage chest, tool kit, sea anemone, \\
 Border Terrier, menu, picket fence, forklift, \\
 yellow ladys slipper, chameleon, dragonfly,  \\
 European garden spider, Airedale Terrier, \\
 frilled-necked lizard, black stork, valley, \\
radio telescope, leopard, crossword, Australian Terrier,\\
Shih Tzu, husky, can opener, assault rifle,\\
fountain pen, harvestman, parallel bars,  half-track,\\
snoek fish, pencil sharpener, submarine, muzzle, \\
eastern diamondback rattlesnake, Miniature Schnauzer,\\
king penguin,canoe, red-breasted merganser,\\
trolleybus, quail, poke bonnet,\\
King Charles Spaniel,  Malinois, \\
solar thermal collector,bucket, dung beetle, \\
Asian elephant, window screen,Flat-Coated Retriever,\\
steel drum, snowplow, handkerchief,\\
tailed frog, church, Chesapeake Bay Retriever,\\
Christmas stocking, sidewinder rattlesnake,\\
oscilloscope, worm snake, eel, \\
wok,  platypus,  alligator lizard,\\
consomme, African rock python \\
\end{tabular}
\\
\hline
Test  & Same as train & 
\begin{tabular}[c]{@{}c@{}}
laptop, bass, joshua\_tree, Leopards, \\
ibis, hedgehog, minaret, crocodile, \\
flamingo, headphone, ketch, saxophone, \\
euphonium, dragonfly, wrench, llama, \\
lamp, kangaroo, yin\_yang, sea\_horse
\end{tabular} 
& Same as train &
\begin{tabular}[c]{@{}c@{}}
hot tub, harmonica, grand piano, hatchet, \\
Pomeranian, missile, slug, vulture, \\
kingsnake, Pembroke Welsh Corgi, cocktail shaker, race car,\\
airship, Tibetan Mastiff, Old English Sheepdog, planetarium,\\
hair clip, Komondor, artichoke, website\\
\end{tabular} 
\\
\hline
\end{tabular}}
\end{center}
\vspace{-5pt}
\caption{Category split for standard and zero-shot experimental settings in N-Caltech101~\cite{ncaltech} and N-ImageNet (Mini)~\cite{kim2021n}. In the standard setting, categories are identical for training and testing. In the zero-shot setting, there is no common category for training and testing.}
\label{tab:data_split}
\vspace{-10pt}
\end{table*}

\section{Implementation Details}

We implement our framework on PyTorch~\cite{paszke2019pytorch} and adopt the Vision Transformer~\cite{vit} with a patch size of 32, called ViT-B/32, as the visual encoder and the transformer~\cite{vaswani2017attention} as the textual encoder. We design the image reconstruction network $\mathcal{G}$ with U-Net~\cite{ronneberger2015u} consisting of two residual blocks~\cite{he2016deep}. For event representation, we utilize the Event Spike Tensor (EST)~\cite{gehrig2019end} using a bin size of 9, which is the same as the original paper. 
For the crop function $\mathcal{H}$ in Sec.~\textcolor{red}{3.3} of the main paper, we use a crop size of $128 \times 128$ while we resize the EST to $224 \times 224$ for all datasets.

We use agglomerative clustering for the clustering algorithm in Sec.~\textcolor{red}{3.4} of the main paper.
For the number of clusters, we use $L=3$ for N-Caltech101 and $L=10$ for N-ImageNet~(Mini).
We use LAMB~\cite{lamb} optimizer with a weight decay of $1\times10^{-4}$ and initialize the learning rate to $6\times10^{-3}$. We train the network with a batch size of 32. We set the weights $\lambda_1, \lambda_2$, $\lambda_3$ and parameter $K$ in PPI as 1, 0.01, 1, and 6, respectively.

\section{Dataset Details}
In Table~\ref{tab:data_split}, we present the details about the category split of each dataset used in our experiments.
In the main paper, the experiments involve two settings: standard and zero-shot.
In the standard setting, the categories are identical for training and testing, evaluating the ability to recognize objects belonging to the same set of categories for training.
In contrast, there is no common category for training and testing in the zero-shot setting.
This evaluates the model's ability to generalize to novel categories that it has not been explicitly trained on.

\section{Additional Experiments and Discussions}
\subsection{Using Ground Truth instead of Pseudo Label}
\begin{table}[h]
\vspace{-12pt}
\centering
\resizebox{0.41\textwidth}{!}{
\begin{tabular}{ccc}
\hline
Method & N-Caltech101 & N-ImageNet (Mini)\\ \hline
Ours   & 82.46 & 30.16 \\
Oracle & \textbf{84.28} & \textbf{32.98} \\
\hline
\end{tabular}}
\vspace{5pt}
\caption{The results of our method using ground truth.}
\vspace{-3pt}
\label{table:oracle}
\end{table}

Table~\ref{table:oracle} demonstrates that using ground truth in training indeed yields better performance than using pseudo labels. Nonetheless, through reliable sampling, our method achieves comparable performance, indicating the superiority of our approach.

\subsection{Ablation Study on Repulsion Loss}
\begin{table}[h]
\setlength{\tabcolsep}{4pt}
\vspace{-10pt}
\centering
\resizebox{0.47\textwidth}{!}{
\begin{tabular}{ccc}
\hline
Loss type & N-Caltech101 & N-ImageNet (Mini) \\ \hline
Category-agnostic~(Eq.~\textcolor{red}{5}) & \textbf{82.46} & \textbf{30.16} \\
Category-aware  & 81.48 & 28.54 \\
\hline
\vspace{1pt}
\end{tabular}}
\caption{The results of using category-agnostic and category-aware repulsion losses.}
\label{table:quan_results}
\end{table}

In Table~\ref{table:quan_results}, we report the result using category-aware repulsion loss that separates only features belonging to the different categories.
The result shows that category-agnostic repulsion is more effective.

\subsection{Experiments with More Prompts}
\begin{table}[h]
\vspace{-10pt}
\centering
\resizebox{0.42\textwidth}{!}{
\begin{tabular}{cc}
\hline
Prompts & N-Caltech101 \\ 
\hline
``image of a [CLASS].'' & \textbf{82.46} \\
``gray image of a [CLASS].'' & 81.30\\
``photo of a [CLASS].'' & 80.64\\
``reconstructed image of a [CLASS].'' & 81.62\\
``clean image of a [CLASS].'' & 81.04 \\
\hline
\end{tabular}
}
\vspace{5pt}
\caption{The results using various text prompts.}
\label{table:text_prompts}
\end{table}
We conduct experiments with five prompt designs. As shown in Table~\ref{table:text_prompts}, the simplest prompt shows the best performance.

\section{Additional Qualitative Results}
\begin{figure*}[t] 
\centering 
\vspace{-2pt}\includegraphics[width=.92\linewidth]{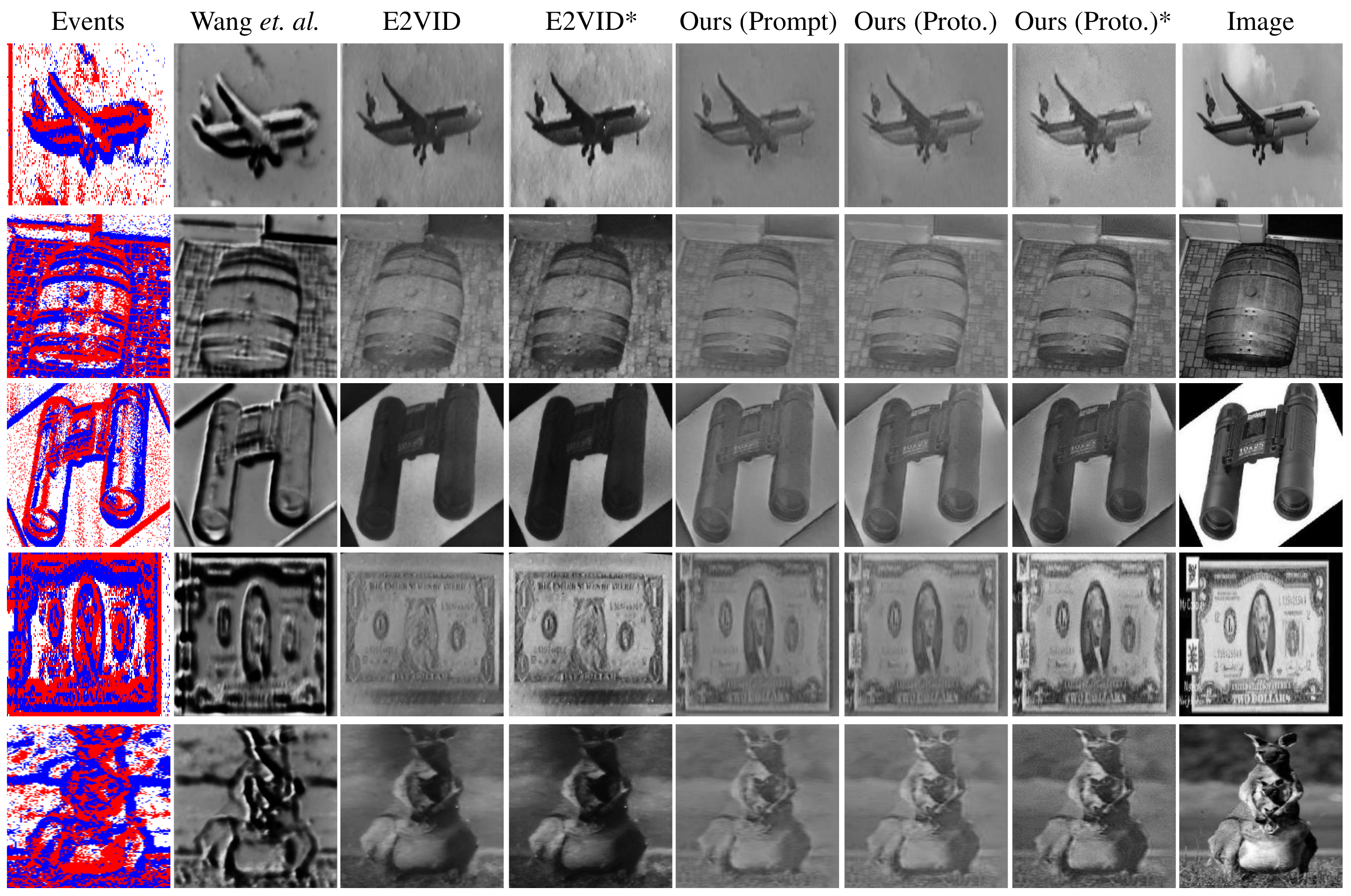} 
\vspace{-5pt}
\caption{Additional qualitative results on N-Caltech101.}
\label{fig:supp_caltech}
\vspace{-5pt}
\end{figure*}

\begin{figure*}[t] 
\centering 
\vspace{-2pt}\includegraphics[width=.92\linewidth]{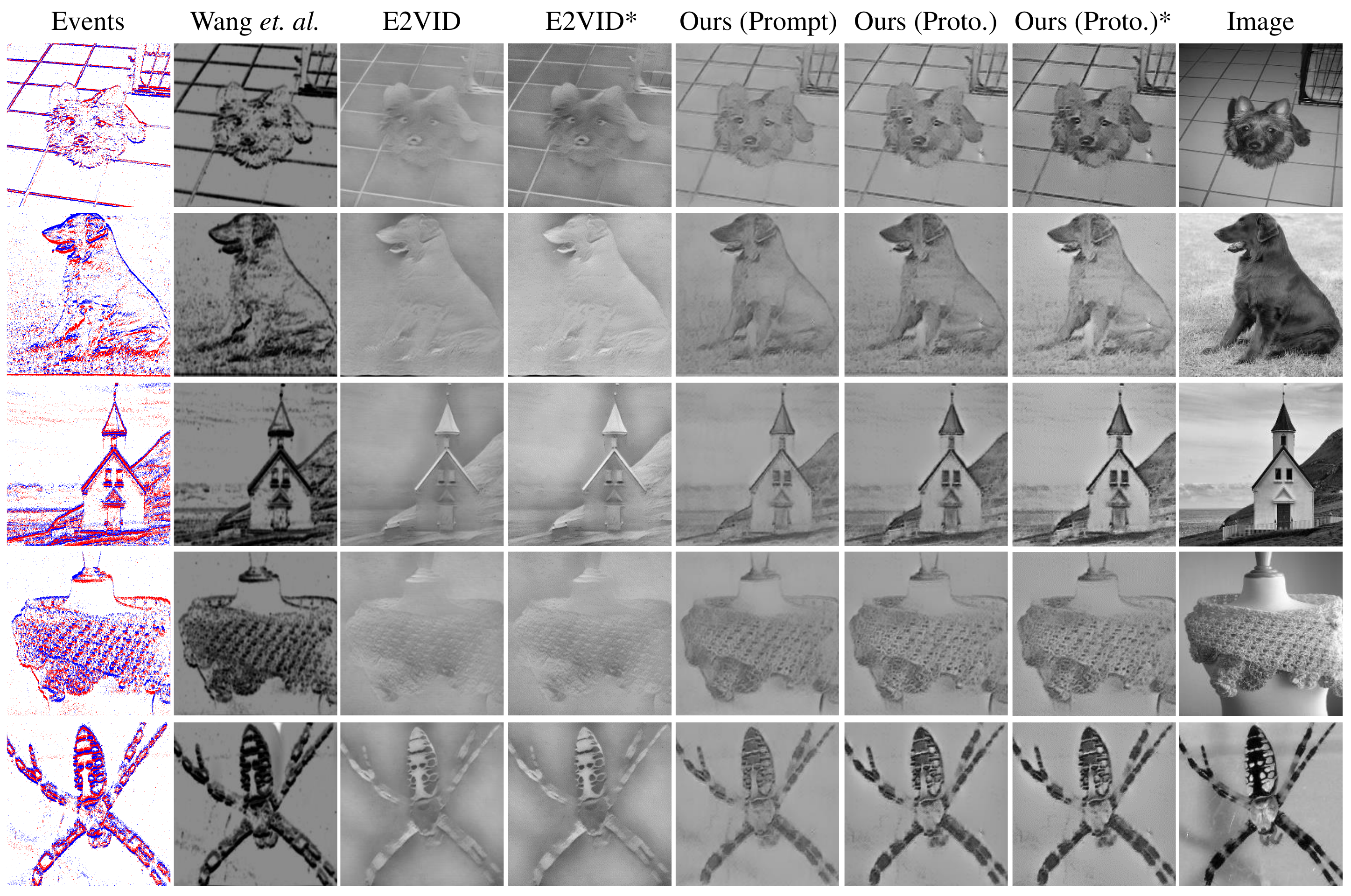} 
\vspace{-5pt}
\caption{Additional qualitative results on N-ImageNet~(Mini).}
\label{fig:supp_imagenet}
\vspace{-5pt}
\end{figure*}

We additionally visualize the qualitative results on N-Caltech101~\cite{ncaltech} and N-ImageNet~(Mini)~\cite{kim2021n} in Fig.~\ref{fig:supp_caltech} and Fig.~\ref{fig:supp_imagenet}, respectively.
Compared to other methods, the proposed method reconstructs images with more fine details.

\end{document}